\documentclass[10pt,twocolumn,letterpaper]{article}

\usepackage{iccv}
\usepackage{times}
\usepackage{epsfig}
\usepackage{graphicx}
\usepackage{amsmath}
\usepackage{amssymb}
\usepackage{bbm}
\usepackage{subfigure}
\usepackage{caption}
\usepackage{algorithm}
\usepackage{algpseudocode}
\DeclareMathOperator*{\argmax}{argmax}

\usepackage[breaklinks=true,bookmarks=false]{hyperref}

\iccvfinalcopy 

\def\iccvPaperID{3273} 

\ificcvfinal\fi

\begin{document}

\title{Ask\&Confirm: Active Detail Enriching for Cross-Modal Retrieval with Partial Query}

\author{Guanyu Cai$^{1,2}$\thanks{Work done during internship at Youtu Lab}, Jun Zhang$^2$, Xinyang Jiang$^{3}$\thanks{Corresponding author: Xinyang Jiang, Xing Sun}, Yifei Gong$^2$,\\ Lianghua He$^1$, Fufu Yu$^2$, Pai Peng$^2$, Xiaowei Guo$^2$, Feiyue Huang$^2$, Xing Sun$^{2\dagger}$\\
Tongji University$^1$, Tencent Youtu Lab$^2$, Microsoft Research$^3$\\
{\tt\small \{caiguanyu,Helianghua\}@tongji.edu.cn,xinyangjiang@microsoft.com,pengpai\_sh@163.com}\\ {\tt\small\{bobbyjzhang,yifeigong,fufuyu,scorpioguo,garyhuang,winfredsun\}@tencent.com}
}

\maketitle
\ificcvfinal\thispagestyle{empty}\fi

\begin{abstract}
Text-based image retrieval has seen considerable progress in recent years. However, the performance of existing methods suffers in real life since the user is likely to provide an incomplete description of an image, which often leads to results filled with false positives that fit the incomplete description. 
In this work, we introduce the partial-query problem and extensively analyze its influence on text-based image retrieval. 
Previous interactive methods tackle the problem by passively receiving users' feedback to supplement the incomplete query iteratively, which is time-consuming and requires heavy user effort. 
Instead, we propose a novel retrieval framework that conducts the interactive process in an Ask-and-Confirm fashion, where AI actively searches for  discriminative details missing in the current query, and users only need to confirm AI's proposal.
Specifically, we propose an object-based interaction to make the interactive retrieval more user-friendly and present a reinforcement-learning-based policy to search for discriminative objects.
Furthermore, since fully-supervised training is often infeasible due to the difficulty of obtaining human-machine dialog data, we present a weakly-supervised training strategy that needs no human-annotated dialogs other than a text-image dataset. 
Experiments show that our framework significantly improves the performance of text-based image retrieval. Code is avaiable at \url{https://github.com/CuthbertCai/Ask-Confirm}.
\end{abstract}
\vspace{-5mm}
\section{Introduction}

Recently, cross-modal retrieval, especially text-based image retrieval has gained increasing attention~\cite{zhang2018deep}. 
Although significant improvement has been achieved with existing methods~\cite{lee2018stacked,zhang2018deep,guo2018dialog} for text-based retrieval, we found in practice their retrieval result is barely satisfactory when users only describe some local regions in an image.

\begin{figure}
    \centering 
    \includegraphics[width=0.78\linewidth]{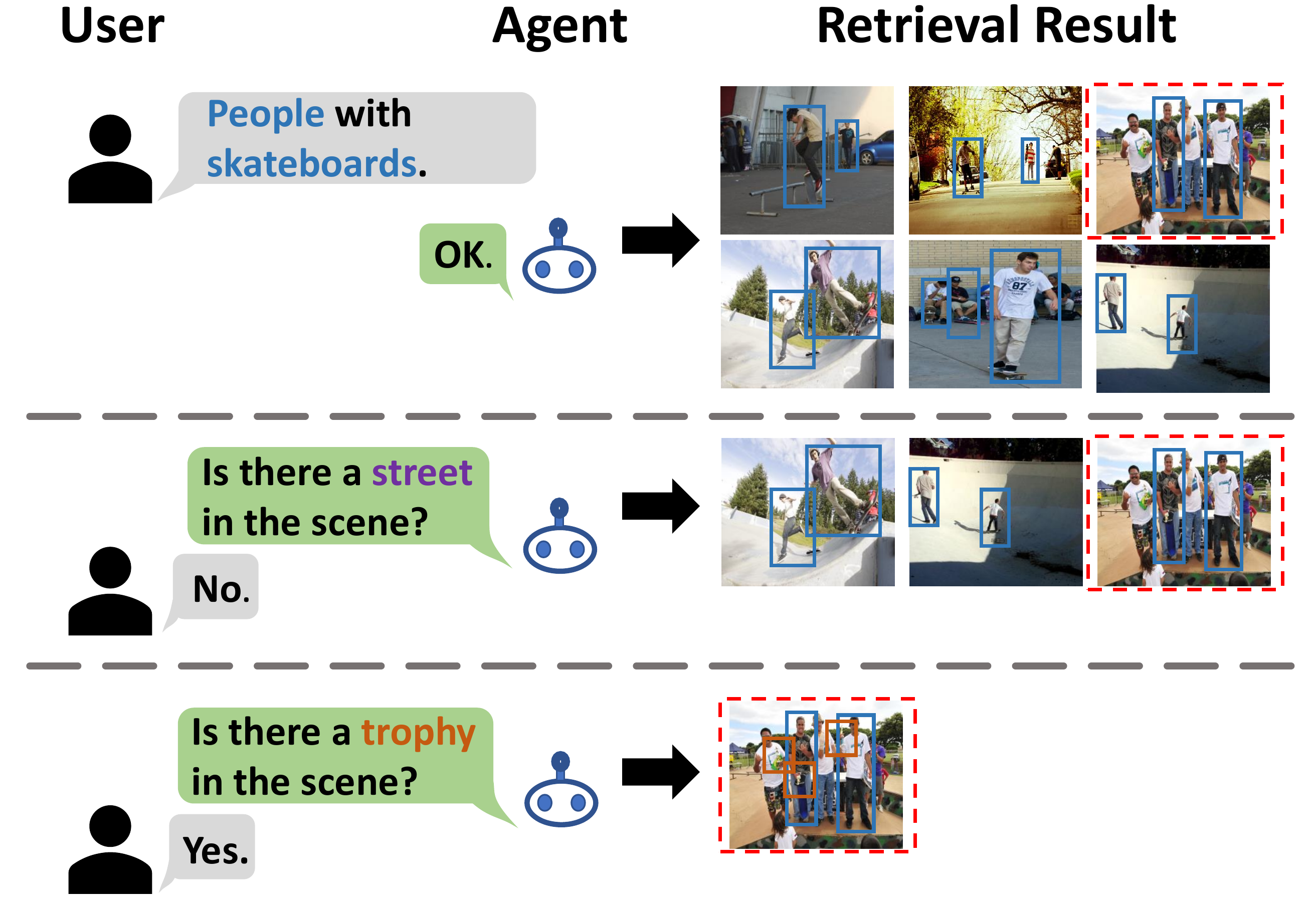}
    \vspace{-3mm}
    \caption{An illustration of Ask\&Confirm. 
    The agent enriches the textual query and narrows down the retrieval scope by iteratively asking users to confirm more information. 
    The target image is highlighted with a red rectangle.}\vspace{-7mm}
    \label{fig:illustration}
\end{figure}

\begin{figure}
    \vspace{-3mm}
    \subfigure[]{
        \begin{minipage}[t]{0.45\linewidth}
        \centering
        \includegraphics[width=1.5in,height=1.in]{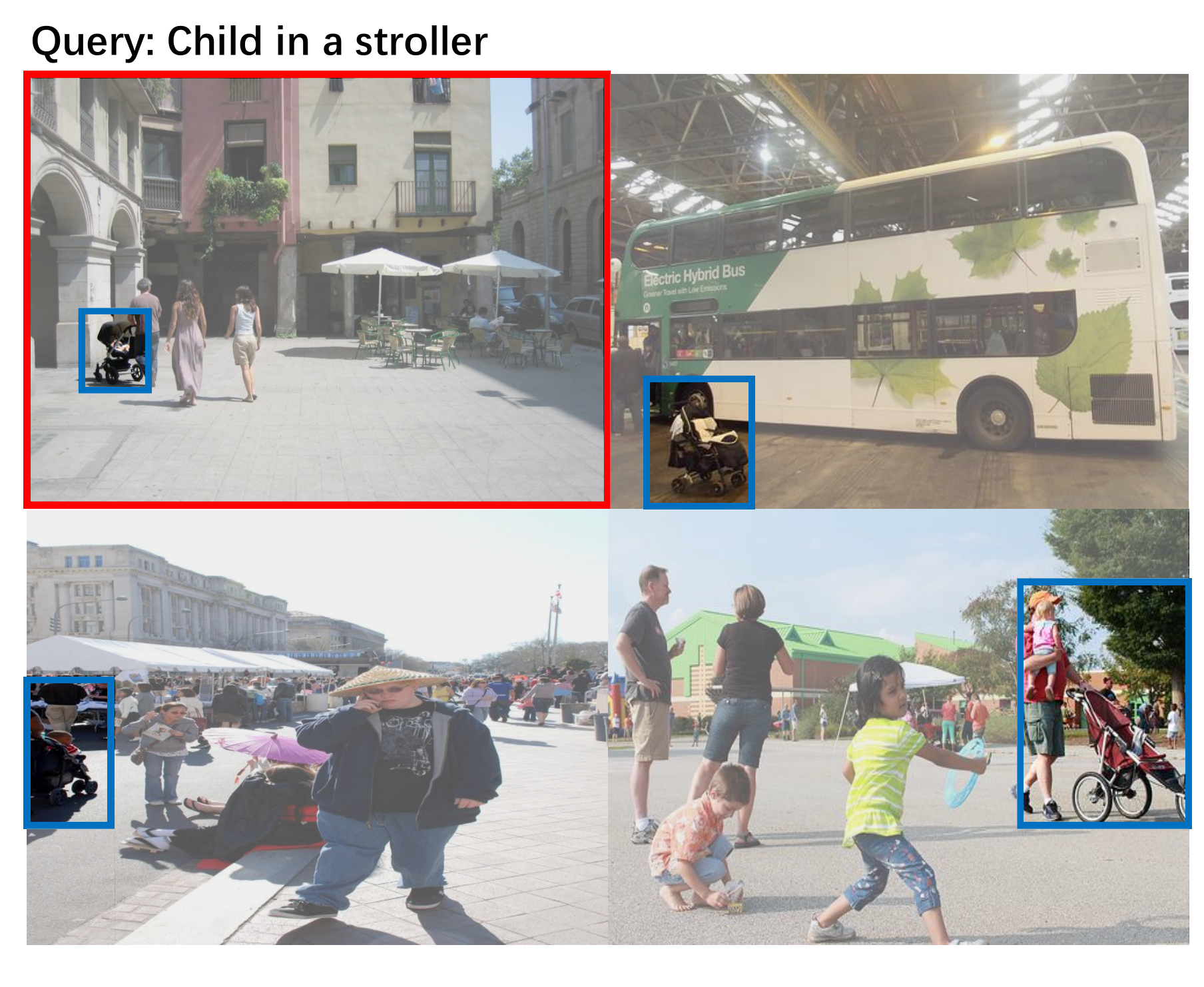}
        \vspace{-10mm}
        \caption*{}
        \end{minipage}
    }\vspace{-2mm}
    \subfigure[]{
        \begin{minipage}[t]{0.45\linewidth}
        \centering
        \includegraphics[width=1.5in,height=1.in]{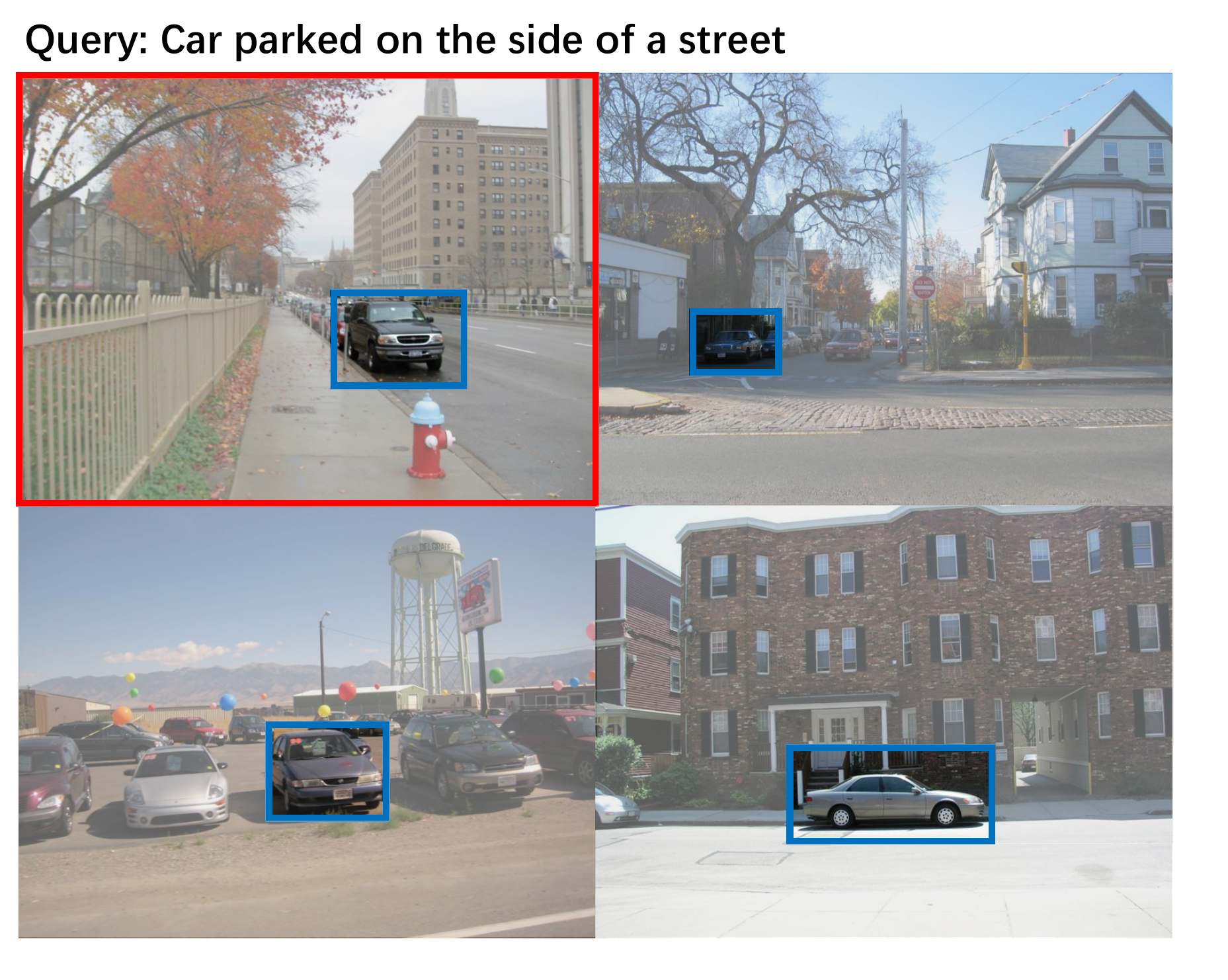}
        \vspace{-10mm}
        \caption*{}
        \end{minipage}
    }\vspace{-2mm}
    \subfigure[R@10]{
        \begin{minipage}[t]{0.45\linewidth}
        \centering
        \includegraphics[width=1.651in]{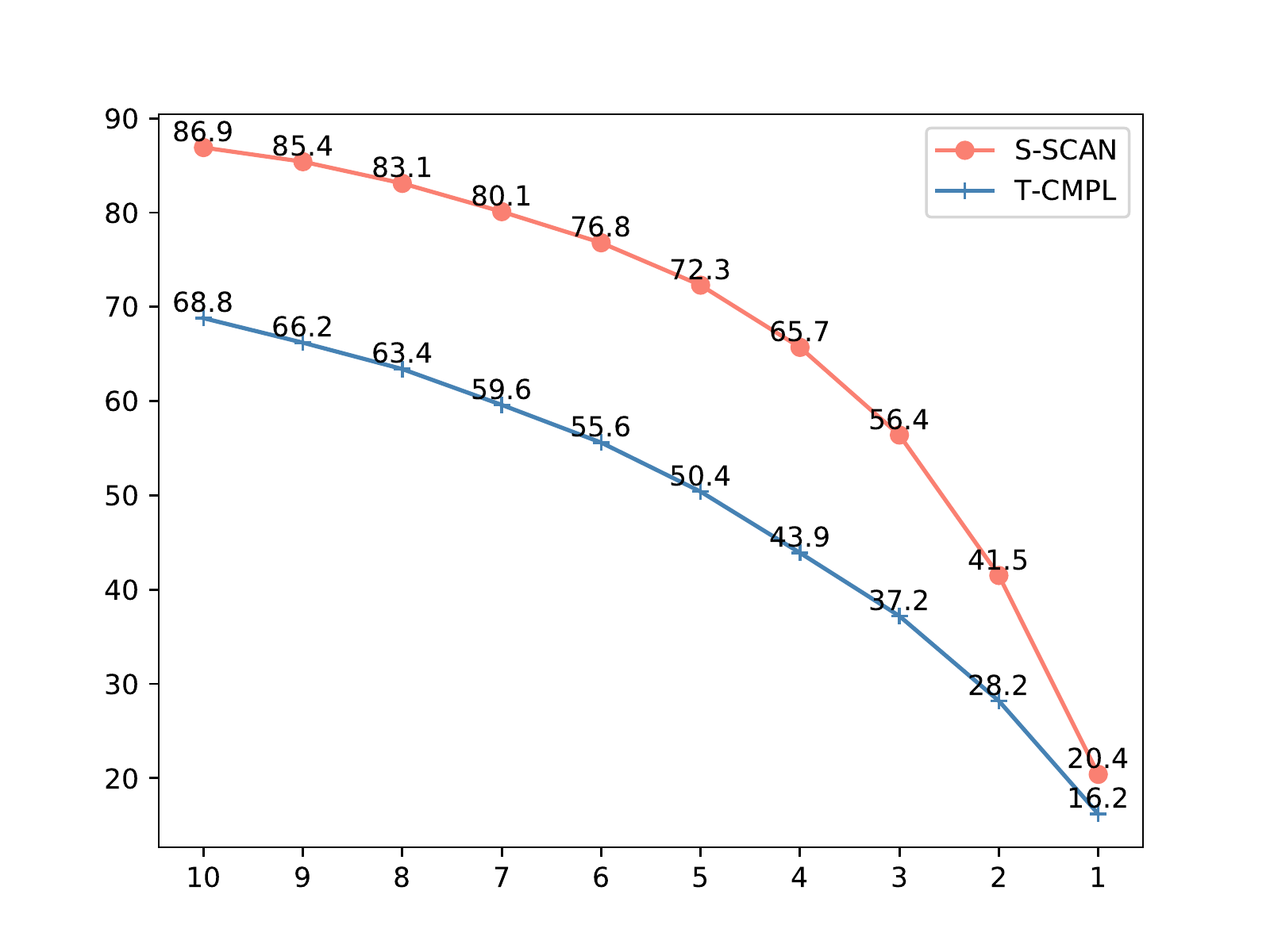}
        \vspace{-15mm}
        \caption*{}
        \end{minipage}
    }
    \subfigure[Mean Rank]{
        \begin{minipage}[t]{0.45\linewidth}
        \centering
        \includegraphics[width=1.651in]{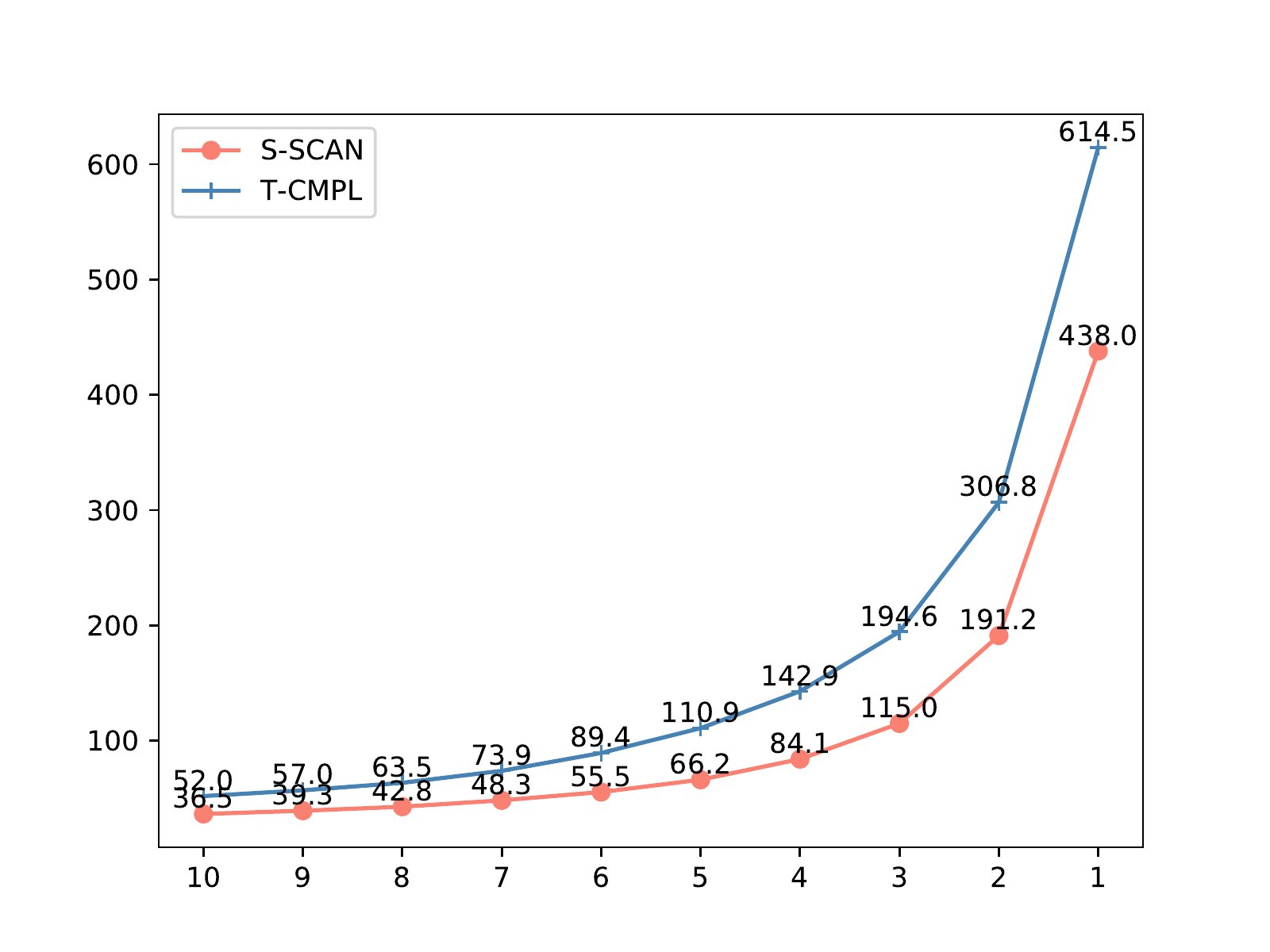}
        \vspace{-15mm}
        \caption*{}
        \end{minipage}
    }
    \vspace{-6mm}
    \caption{Effect of partial queries. (a) and (b) are visualizations of partial-query retrieval. The target image is surrounded by a red box and the others are the top three ranked scenes. The region that matches the query is surrounded by a blue box. (c) and (d) demonstrates R@10 and Mean Rank of a retrieval model as queries decrease. The horizontal axis represents the query number.}
    \vspace{-7mm}
    \label{fig:partial_effect}
\end{figure}

In this work, we introduce a new concept of {\it partial-query} problem in text-based image retrieval, where the initial text query only describes some objects in the target image. 
Studies~\cite{van2003selective,shomstein2004control} have found that when examining an image, people tend to only focus on the objects that stand out the most. 
This could lead to problems where the objects that people focus on are not the discriminative objects that can distinguish the target image from similar candidates, thus making the user's input insufficient for retrieving the target image. As shown in Figure~\ref{fig:partial_effect} (a) and (b), a cross-modal retrieval model performs poorly when a query is only partially given. In both examples, the target image ranks lower than 1000 th, while the other false positives rank top three. A common object (blue box) described by the partial query is presented in all images. However, the rest of images are vastly different. For example, in Figure~\ref{fig:partial_effect} (a), besides the stroller mentioned in the query, the target image consists of umbrellas, chairs, and so on. Whereas the others consist of different objects like trees and buses. 
If the retrieval model receives a complete description including all objects, existing methods~\cite{lee2018stacked,liu2019focus,ijcai2019-526} perform excellently. To show how the partial query hurts retrieval, we test two text-image retrieval models, S-SCAN and T-CMPL on Visual Genome~\cite{krishna2017visual}, which are modified from SCAN~\cite{lee2018stacked} and CMPL~\cite{zhang2018deep}. The implementation is detailed in Section~\ref{exp}. For each image, its complete description includes 10 captions for different regions. We gradually decrease the number of captions and use them as queries to retrieve the target image. As shown in Figure~\ref{fig:partial_effect} (c) and (d), for both models, R@10 decreases and Mean Rank increases as the degree of incompletion increases. These results reveal that partial queries should be tackled for a robust retrieval model.

Existing interactive retrieval models~\cite{flickner1995query,wu2004willhunter,rui1998relevance,kovashka2012whittlesearch,kovashka2013attribute,yu2017fine,guo2018dialog,vo2019composing} tackle the partial query by involving feedback of users in the retrieval process. Given the initial queries from users, these methods first give several relevant candidates that could potentially be the target image. By comparing the target image with these reference images, users give the retrieval method different forms of feedback to describe the difference between them, such as scores~\cite{rui1998relevance,wu2004willhunter}, tags~\cite{kovashka2012whittlesearch,kovashka2013attribute,kovashka2017attributes} or descriptions~\cite{guo2018dialog,tan2019drill}. 
The models then refine the retrieval results according to the user feedback and continue next round of iteration until the target image is found. 
Previous methods only passively receive additional information from users, so users need to have substantial practice and expert knowledge on the retrieval system to give discriminative feedback that can quickly narrow down the retrieval range. 
Hence, to free users from the burden of analyzing the retrieval results and looking for the discriminative information, we propose that the retrieval model itself should be able to actively search for the  discriminative information the current query misses. 
Another problem of previous interactive retrieval models is time-consuming. For example, description-based methods~\cite{guo2018dialog,tan2019drill} require users to input long sentence feedback and tag-based methods~\cite{kovashka2012whittlesearch,kovashka2013attribute,kovashka2017attributes} require users to input a bunch of attributes. Hence, we propose a framework where users only need to make simple yes/no confirmation on AI's question.

In this paper, we propose a novel interactive retrieval framework called {\it Ask\&Confirm} as shown in Figure~\ref{fig:illustration}. The agent first retrieves a set of relevant candidates from the gallery based on initial text queries. Then, it will analyze the retrieval results and the overall status of gallery, and actively select discriminative object candidates for users to confirm their presence. Based on users' confirmation, the agent narrows down the range of candidates and eventually gathers enough information to locate the target image. 
Instead of passively receiving user feedback, a reinforcement learning (RL) based policy is trained to actively search for the discriminative objects missed in the query, and use these objects to distinguish the target image from the rest of gallery. 
In this active object-based interaction, users only need to confirm the existence of the proposed objects in the target image, no expert knowledge on the retrieval task and extra effort is needed. 
Moreover, unlike previous RL based interactive methods~\cite{guo2018dialog,maeoki2020interactive} that require human-annotated dialogs which is impractical to widely collect, our Ask\&Confirm framework is trained in a weakly-supervised manner, where only text-image pairs are needed.

The contributions of our framework are as follows: 1) To our knowledge, this is the first work that formally addresses and analyzes the problem of partial query in cross-modal retrieval. 
2) Instead of passively receiving missing details from user feedback, we propose a novel interactive retrieval framework {\it Ask\&Confirm} that introduces an active object-based interaction to actively select the most discriminative objects for users to confirm.  
3) Rather than using human-annotated dialogs, we propose a weakly-supervised reinforcement learning framework to optimize the interactive policy that explores the statistical characteristics of the gallery. 
Experiments show that our framework is effective and robust with partial queries.

\section{Related Work}

\subsection{Text-based Image Retrieval}

Most text-based image retrieval approaches are based on deep neural networks~\cite{zhang2018deep,lee2018stacked,liu2019focus,ji2019saliency,ijcai2019-526,faghri2018vse++}. 
The objective of them is to accurately measure the similarity between the inputs from two different modalities. Cross-Modal Projection Learning (CMPL)~\cite{zhang2018deep} is proposed to pull image and text embeddings into an aligned space. To further enhance the retrieval in a fine-grained way, ~\cite{lee2018stacked,liu2019focus,ji2019saliency,ijcai2019-526} proposed different attention-based approaches, applying visual attention between every image region and word. 

\subsection{Query Expansion}

Query expansion tackles incomplete information. Different from partial queries that are complete sentences of local regions, it focuses on queries that are incomplete sentences. An incomplete sentence as the query leads to poor retrieval. Thus, query expansion methods are proposed~\cite{4438522,liu2008query,natsev2007semantic,de2016knowledge,he2016framework}.~\cite{4438522} learns users' searching history to generate expansion.~\cite{liu2008query} explores expansion by calculating similarity distance in thesaurus indexed collections. Other methods~\cite{natsev2007semantic,de2016knowledge,he2016framework} that focus on image or video retrieval provide expansion based on knowledge bases. 

 
\subsection{Visual Dialog}
Visual dialog aims to let the machine understand the visual content and have a natural conversation with the user about it. After examining the image, the agent can answer the user's questions on different aspects. Mainstream approaches are based on policy-based reinforcement learning to achieve good question-answer performance~\cite{padmakumar2020dialog,das2017visual,das2017learning}. However, the dialogs are purely text-based for both the questioner and answer agent, and a manually annotated dialog dataset is needed to train a visual dialog system. 

\subsection{Interactive Image Retrieval}

The retrieval model is hard to locate the target image with the initial query. Inspired by visual dialog, interactive image retrieval systems~\cite{wu2004willhunter,rui1998relevance,kovashka2012whittlesearch,kovashka2017attributes,kovashka2013attribute,parikh2011relative,liao2018knowledge,DBLP:conf/bmvc/Murrugarra-Llerena18,MURRUGARRALLERENA2021103204} are proposed to solve this problem. In these systems, users give feedback to an agent according to a reference image. There are two types of feedback: relevance and difference. For the former one~\cite{wu2004willhunter,rui1998relevance}, users give relevance scores for the current retrieval results. Then the system re-ranks its retrieval results by using the user's feedback. For the latter one~\cite{kovashka2012whittlesearch,kovashka2017attributes,kovashka2013attribute,parikh2011relative,yang2020glance}, users tell the difference between the target image and a reference image to the system with tags or descriptions. The system then whittles away the irrelevant images and ranks the correct one to the top.

\section{Method}

\subsection{Object-based Interaction}
\label{object}


In a partial-query problem, one of the most important tasks for an interactive retrieval model is to obtain the missing discriminative information that can distinguish the target image from others.  
Generally, the demands of more discriminative information and less user effort are contradictory, because more information usually means that the user has to pay more effort to think about what is the most discriminative thing and to input more descriptions. For example, tag-based methods~\cite{kovashka2012whittlesearch,kovashka2013attribute,kovashka2017attributes} only require the user to point out a different attribute between the target image and a reference image, but they hardly filter out many negative images per round because too little discriminative information is provided. On the contrary, description-based methods~\cite{guo2018dialog,tan2019drill} require the user to give long sentence feedback that enriches more details but pays more user effort. 

In Ask\&Confirm, we propose an object-based interaction where a RL-based policy actively searches for discriminative object candidates for users to confirm, then users just need to confirm whether objects are in the target image. Under the auxiliary of the active policy, the demands of more discriminative information and less user effort are simultaneously satisfied. 

We choose object-based interaction based on two main reasons: (1) objects in an image are discriminative enough to distinguish different images, (2) objects can be easily obtained with a pre-trained detector such as RCNN~\cite{anderson2018bottom}. 

Firstly, we discover that the distribution of objects in an image gallery is generally low-entropy, making it a discriminative feature for retrieving the target image. For example, in Visual Genome~\cite{krishna2017visual}, some objects, such as ``trophy" and ``skateboard", rarely appear. If an image includes them, they are discriminative enough to narrow down the retrieval scope quickly.
To verify this observation, 
two types of queries are compared using the same retrieval method S-SCAN : partial query only and supplement partial query with the name of the objects. 
As shown in Table~\ref{tab:gt_attribute}, remarkable improvements are achieved by adding object words, verifying that objects contain discriminative information to distinguish the target image from the rest of the gallery. 

Second, the convenience of obtaining objects of an image also makes the object-based interaction practical. Previous text-based image retrieval methods~\cite{faghri2018vse++,lee2018stacked,liu2019focus,ijcai2019-526} extract image features by an object detector~\cite{anderson2018bottom}. By reusing the detector, we can directly obtain objects of each image.

\begin{table}
\begin{center}
\begin{tabular}{|l|c|c|c|c|}
\hline
Method & R@1 & R@5 & R@10 & MR\\
\hline\hline
S-SCAN& 4.5 & 13.6 & 20.4 & 416.0 \\
S-SCAN+Objects& 46.4 & 70.2 & 78.4 & 28.4 \\
\hline
\end{tabular}
\end{center}
\vspace{-5mm}
\caption{Retrieval improvements over S-SCAN with ground-truth object descriptions. MR means Mean Rank.}
\label{tab:gt_attribute}
\vspace{-6mm}
\end{table}

\subsection{Interactive Retrieval Agent}
\label{interaction}

By adopting the proposed object-based interaction, we propose an interactive retrieval agent to tackle the partial-query problem. It takes the charge of extracting features, interacting with the user and retrieving the target image. In this section, we illustrate how the agent works, especially how it actively searches for object candidates for the user to confirm, which greatly reduces the user effort.

Define a set of captions $Q=\{q_n\}^{N_Q}_{n=1}$ that composes descriptions of an image $i$, where each $q_n$ describes a region. By regarding $Q$ as queries, the goal of a retrieval agent $R$ is to retrieve the target image $i_*$ from a gallery $I=\{i_n\}^{N}_{n=1}$ through $T$ rounds interaction with the user. The partial-query problem considers that $Q$ only describes parts of an image instead of the full image.

\begin{figure*}
    \centering
    \vspace{-4mm}
    \includegraphics[width=0.70\textwidth]{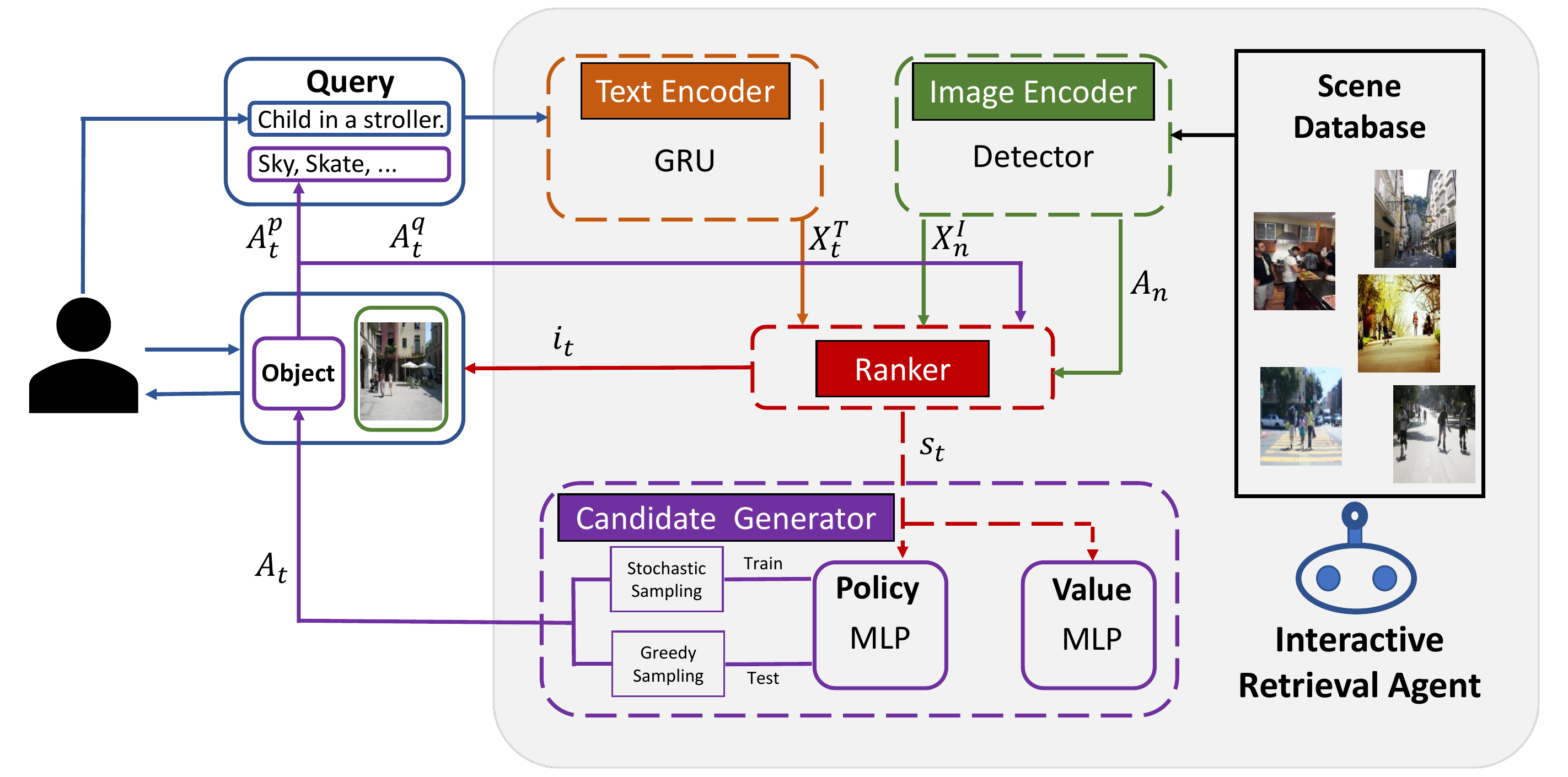}
    \vspace{-4mm}
    \caption{The proposed interactive cross-modal retrieval framework of Ask\&Confirm. The interactive retrieval agent gradually enriches details of an image by heuristically providing users with object candidates.}
    \label{fig:framework}
    \vspace{-6mm}
\end{figure*}

The interactive retrieval agent $R$ includes four main components: {\bf Text Encoder}, {\bf Image Encoder}, {\bf Candidate Generator} and {\bf Ranker}. As the interactive workflow~\ref{fig:framework} illustrated, Text Encoder and Image Encoder embed partial queries and images to a textual-visual feature space as textual features and visual features respectively. At each round, Candidate Generator actively searches for the most discriminative objects as candidates for the user to confirm. Given the objects, the user confirms them as the positive or negative, where positive objects refer to the ones that exist in the target image, vise versa. Then, names of positive objects are added to the partial query and the new query's feature is updated by the Text Encoder.
Finally, based on positive objects, negative objects and the features of queries and images, Ranker retrieves the target image. 
In detail, Ranker first computes an initial similarity between the textual query and visual features. 
Secondly, the initial similarity is further refined by the user-confirmed objects. 
If a gallery image contains the negative objects, the similarity between the image and queries would be refined to a lower value. 
The  retrieval result of the current round is given by the refined similarity. 
Below we provide details on the specific design of each component.

{\bf Text Encoder.} At $t$ th round, the input partial queries are denoted as $Q_t=\{q_n\}^{N^t_Q}_{n=1}$. They are embedded into textual features by Text Encoder (TE):
\begin{equation}
\vspace{-1mm}
    x^T_n=TE(q_n), q_n\in Q_t
\end{equation}
where $x^T_n$ denotes a texture feature. The set of all textual features of $Q_t$ is denoted as $X^T_t=\{x^T_n\}^{N^t_Q}_{n=1}$. In detail, we use a gated recurrent unit as $TE$ just like~\cite{tan2019drill}.

{\bf Image Encoder.} Given an image gallery $I=\{i_n\}_{n=1}^{N}$, Image Encoder (IE) extracts the visual feature and detects objects for each image:
\begin{equation}
    \vspace{-1mm}
    (x^I_n, A_n)=IE(i_n)
\end{equation}
where $x^I_n$ denotes the visual feature of $i_n$ and $A_n$ denotes the objects $\{a_1, a_2, ...\}$ that appear in $i_n$. The set of all visual features of $I$ is denoted as $X^I=\{x^I_n\}^{N}_{n=1}$.

{\bf Candidate Generator.} At $t$ th round, Candidate Generator actively searches for the most discriminative objects as candidates for the user to confirm positive objects that appear in the target image $i_*$. These candidates are denoted as $A_t=\{a_n\}^{N_A}_{n=1}$. The user confirms positive objects $A^p_t=\{a^p_n\}^{N^p_A}_{n=1}$, thus, the rest of $A_t$ are negative objects. They are denoted as $A^q_t=\{a^q_n\}^{N^q_A}_{n=1}$ where $N^q_A+N^p_A=N_A$.

The text of $A^p_t$ is used as the additional description of $i_*$. It is denoted as $Q^c_t=\{\mathbb{T}(a^p_n)\}^{N^p_A}_{n=1}$, where $\mathbb{T}(a^p_n)$ is the word of $a^p_n$. To enrich details of the target image, the additional description is added into queries where $Q_t=Q_{t-1}\cup Q^c_t$.

{\bf Ranker.} Given $X^T_t$, $X^I$, and $A^q_t$, Ranker gives a retrieval result of $t$ th round. Firstly, Ranker computes the similarity between queries and each image, where $S_{t,n}(X^T_t,x^I_n)$ denotes the similarity between $X^T_t$ and $x^I_n$. Secondly, if $i_n$ contains negative objects belong to $A^q_t$, we refine $S_{t,n}$ with a lower value where $S_{t,n}:=S_{t,n}\times0.9$. With the refined similarity, Ranker gives a retrieval result.

\subsection{Weakly-supervised Policy Learning}

The key of Ask\&Confirm to satisfy the demand of more discriminative information and less user effort is an active search policy. It selects the most discriminative objects as candidates for users to confirm, according to the textual feature and the object distribution of an image gallery. Thus, it frees the user to think about what are the most discriminative objects and input long sentence feedback.

In this work, the active search policy is learned with a weakly-supervised RL-based training. The weakly-supervised policy learning automatically finds an optimal policy by letting the agent iteratively interact with users and self-update based on the users' feedback. The whole policy learning is very concise and can be easily conducted in a weakly-supervised manner, because we only need to know objects in each image and users' feedback can be mimicked by ground-truth objects in the target image. 
We can even just reuse the detector for extracting image features to detect objects. On the contrary, previous dialog-based retrieval methods~\cite{guo2018dialog,maeoki2020interactive,das2017learning} require burdensome collections of chatting sessions. The superiority that our method needs no extra data collections makes it more practical.


{\bf Reinforcement Learning.} 
The policy obtained by Candidate Generator is modeled as 
 a policy net $\pi$, parameterized with $\phi_{\pi}$, which outputs each object's probability $P(a)$ of getting selected.  
The five components in our policy learning {\it action}, {\it state}, {\it policy}, {\it value} and {\it reward} are as follows: 

{\it Actions} refers to the objects selected by Candidate Generator at each round, i.e., $a \in \mathcal{A}$, $\mathcal{A}$ is the set of all objects. 

{\it State} $s^t$ is defined as a concatenation of  $s_1^t=\sum_{n=1}^{N_Q^t}x_n^T/N_Q^t$ and $s_2^t=P_r(a)$, where $P_r(a)$ is the distribution of $a$ among the top 100 images generated by the Ranker. We utilize such design to make $\pi$ aware of information both from partial queries and the ranking list. 

{\it Reward} is defined as the similarity between textual features and the target visual feature, i.e., $S(X_t^T, x_*^I)$ where $x_*^I$ is the visual feature of the target image. 

{\it Policy} $\pi$ is implemented with a three-layer MLP. The object sampling distribution $P(a)$ is approximated with $\pi(s^t)$. 

{\it Value} is estimated with $V(s^T)$. The value net $V$ is implemented with a two-layer MLP, parameterized with $\phi_v$.

Given the actions, state, reward, value and policy, a Proximal Policy Optimization (PPO)~\cite{schulman2017proximal} is applied to optimize the policy net $\phi_\pi$ and $\phi_v$. 
Please refer to the original paper of PPO for more details.

{\bf Shaping.} RL is hard to converge for dialog agents~\cite{ng1999policy,guo2018dialog}, thus, previous RL-based dialog agents~\cite{guo2018dialog,padmakumar2020dialog,das2017learning} adopt a supervised learning with annotated dialogs for shaping the RL training. 
To avoid the burdensome human annotation, we propose a weakly-supervised shaping method without annotated dialogs. 
Our motivation is that objects in the target image should have a high probability to be selected, because adding these objects into the queries could potentially significantly increase the similarity between queries and the target image. 
However, this probability is infeasible to obtain during test time, because the target image is unknown. 
As a result, instead of obtaining the probability of objects existing in the target image, we approximate it with the probability of objects that semantically relevant to the target image's corresponding query. 
For example, if the target image's corresponding query is ``a man is surfing", we can infer that objects relevant to this query (e.g.,  ``man", ``sea" and ``surfboard") should have a high probability to appear in the target image. 
The semantic relevance between an object and a query can be estimated by the conditional probability of an object $a_j$ existing in the query's corresponding target image, given the query $Q_t$, denoted as $P(a_j|Q_t)$. $P(a_j|Q_t)$ can be estimated by computing the frequency of $a_j$ and $Q_t$ both appearing in the same target image $i_k$: 
\begin{equation}
\small
    \vspace{-2mm}
    P(a_j|Q_t) =\frac{\sum_{k=1}^{N}\mathbbm{1}(a_j\in i_k||Q_t\in i_k)}{\sum_{m=1}^{|\mathcal{A}|}\sum_{k=1}^{N}\mathbbm{1}(a_m\in i_k||Q_t\in i_k)}
\end{equation}
where $\mathbbm{1}(\cdot)$ is an indicator function. $Q_t\in i_k$ is a corresponding query of $i_k$ and $a_j\in i_k$ denotes an object in $i_k$. 

A practical problem is that $Q_t$ hardly appears in different $i_k$, which causes $\sum_{k=1}^{N}\mathbbm{1}(a_j\in i_k||Q_t\in i_k)$ always being $1$. Thus, we use a set of words $\{w_n\}_{n=1}^{N_w}$ to represent $Q_t$, where $w_n$ is a tokenized word in $Q_t$. The tokenized word $w_n$ could appear in different images. $\mathbbm{1}(Q_t\in i_k)$ is replaced with $\sum_{n=1}^{N_w}\mathbbm{1}(w_n\in i_k)$. $P(a_j|Q_t)$ is then modified to:
\begin{equation}
\small
    \vspace{-2mm}
    P(a_j|Q_t) =\frac{\sum_{k=1}^{N}\sum_{n=1}^{N_w}\mathbbm{1}(a_j\in i_k||w_n\in i_k)}{\sum_{m=1}^{|\mathcal{A}|}\sum_{k=1}^{N}\sum_{n=1}^{N_w}\mathbbm{1}(a_m\in i_k||w_n\in i_k)}
\end{equation}
Guiding with $P(a|Q_t)$, We then train $\pi$ by optimizing
\begin{equation}
\small
    \vspace{-2mm}
    \mathcal{L}_s=\sum\nolimits_{t=1}^{N_s}(P(a|Q_t)-\pi(s^t))^2
\end{equation}
where $N_s$ means that $\mathcal{L}_s$ is optimized for every $N_s$ rounds.    

Combining RL with the shaping, loss of the policy learning process is $\mathcal{L}=\mathcal{L}_p+\alpha\cdot\mathcal{L}_s$, where $\mathcal{L}_p$ denotes the loss of PPO and coefficient $\alpha$ is used to balance the RL learning and shaping. The shaping is crucial in our method otherwise the training process cannot converge.

\section{Experiments}
\label{exp}

{\bf Dataset.} 
There is no existing benchmark for interactive partial-query retrieval and we build a new dataset based on Visual Genome~\cite{krishna2017visual}. 
In Visual Genome, multiple regions are detected by an object detector~\cite{anderson2018bottom} for each image, and each of the object region is annotated with a description. 
We preprocess the data by following the protocol in~\cite{tan2019drill}, resulting in 105,414 images. 
Images are split into 92,105/5,000/9,896 for training/validation/testing. 
To perform an interactive partial-query retrieval without extra data collection, 
we regard a region caption as a partial query offered by users and objects in the target image as feedback from users. 
All evaluations are performed on the test split. 

{\bf Baselines.} 
Ask\&Confirm is a simple framework compatible to any cross-modal retrieval methods. 
We implement variants of SCAN~\cite{lee2018stacked} and CMPL~\cite{zhang2018deep}, which are named Simplified SCAN (S-SCAN) and CMPL with Triplet loss (T-CMPL) respectively, as the basic retrieval models and build the proposed interactive retrieval agent on them. 
Both of the variants adopt the text and image encoder in Section~\ref{interaction} to obtain textual features $X^T=\{x^T_j\}^{J}_{j=1}$ and visual features $X^I=\{x^I_{k,m}\}^{K,M}_{k,m=1}$. (a) {\bf S-SCAN}: We modify the bidirectional attention mechanism in SCAN to a unidirectional one to adopt multi-query inputs. Thus, the similarity between $x^T_j$ and $x^I_k$ is modified as
\begin{equation}
\small
\vspace{-2mm}
    S_{j,k}(x^T_j,x^I_k)=\frac{1}{M}\sum_{m=1}^{M}\gamma_{j,k}\cdot cos(x^T_j,x^I_{k,m})
\end{equation}
where $\gamma_{j,k}=\frac{exp(cos(x^T_j,x^I_{k,m}))}{\sum_{m=1}^{M}exp(cos(x^T_j,x^I_{k,m}))}$ and $cos$ denotes the cosine similarity. The similarity between $X^T$ and $x^I_k$ is the average of $S_{j,k}$ among all $x^T_j$.
(b) {\bf T-CMPL}: Similar to CMPL, we adopt global alignment to match textual and visual features without any attention mechanisms. Thus, the similarity between $x^T_j$ and $x^I_k$ is
\begin{equation}
\small
\vspace{-2mm}
    S_{j,k}(x^T_j,x^I_k)=cos(x^T_j,\frac{1}{M}\sum^{M}_{m=1}x^I_{k,m})
\end{equation} 
The similarity between $X^T$ and $x^I_k$ is the average of $S_{j,k}$ among all $x^T_j$.

Both S-SCAN and T-CMPL are optimized with a common ranking loss.  
It is clear that Ask\&Confirm focuses on the interactive mode and is independent of the network architecture and similarity computing. Thus, Ask\&Confirm can adopt any existing cross-modal retrieval models.

{\bf Implementation Details.} During training, $T$ is set to 20 to conduct twenty-round interaction. In each round, we set $N_A=10$ which means sampling 10 objects from the object sampling distribution $P(a)$. During testing, we vary $T$ and $N_A$ and apply a greedy sampling to choose objects with the highest probabilities. Similar to~\cite{anderson2018bottom}, we utilize a Faster RCNN pretrained on Visual Genome with 1600 object categories to extract features of the top 36 regions and predict objects of regions. Textual and visual features are mapped into vectors with a dimension of 256. For the optimization of policy learning, we update all parameters for every 600 rounds and adopt Adam~\cite{kingma2014adam} as the optimizer. Learning rates of $\phi_\pi$, $\phi_v$ are $3e^{-4}$ and $1e^{-3}$. Coefficient $\alpha$ is set to $1000$. All models are trained for 500 epochs.

{\bf Evaluation Metrics.} We adopt the common R@K (K=1, 5, 10) metric and Mean Rank (MR) to measure the retrieval performance. R@K indicates the percentage of the queries where at least one ground truth is retrieved among the top-K candidates. 

\begin{figure}
    \vspace{-2mm}
    \subfigure[Query 1/Action 10]{
        \begin{minipage}[t]{1\linewidth}
        \centering
        \includegraphics[width=3.3in]{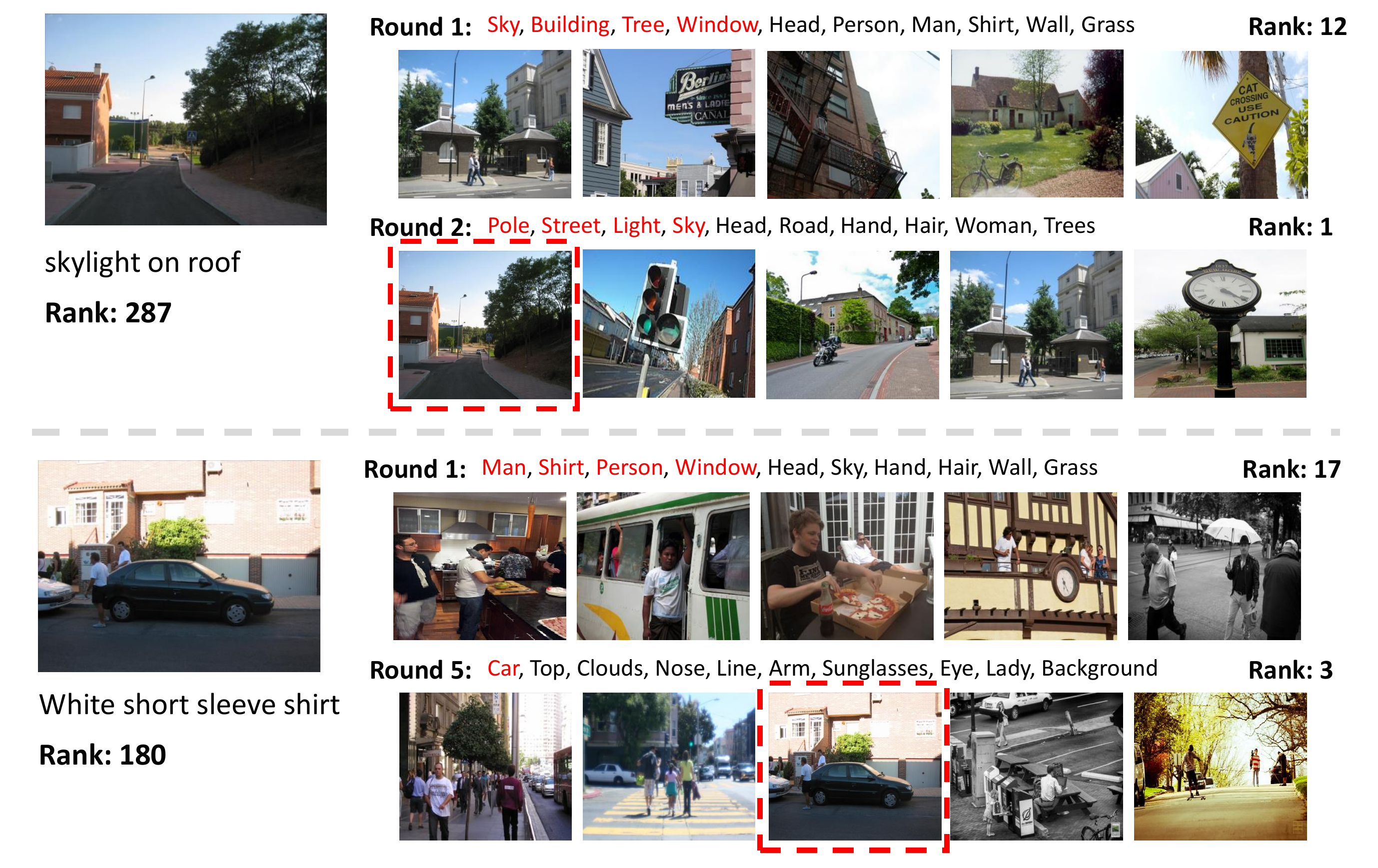}
        \vspace{-16mm}
        \caption*{}
        \end{minipage}
    }\hspace{-15mm}\vspace{-3mm}
    \subfigure[Query 2/Action 5]{
        \begin{minipage}[t]{1\linewidth}
        \centering
        \includegraphics[width=3.3in]{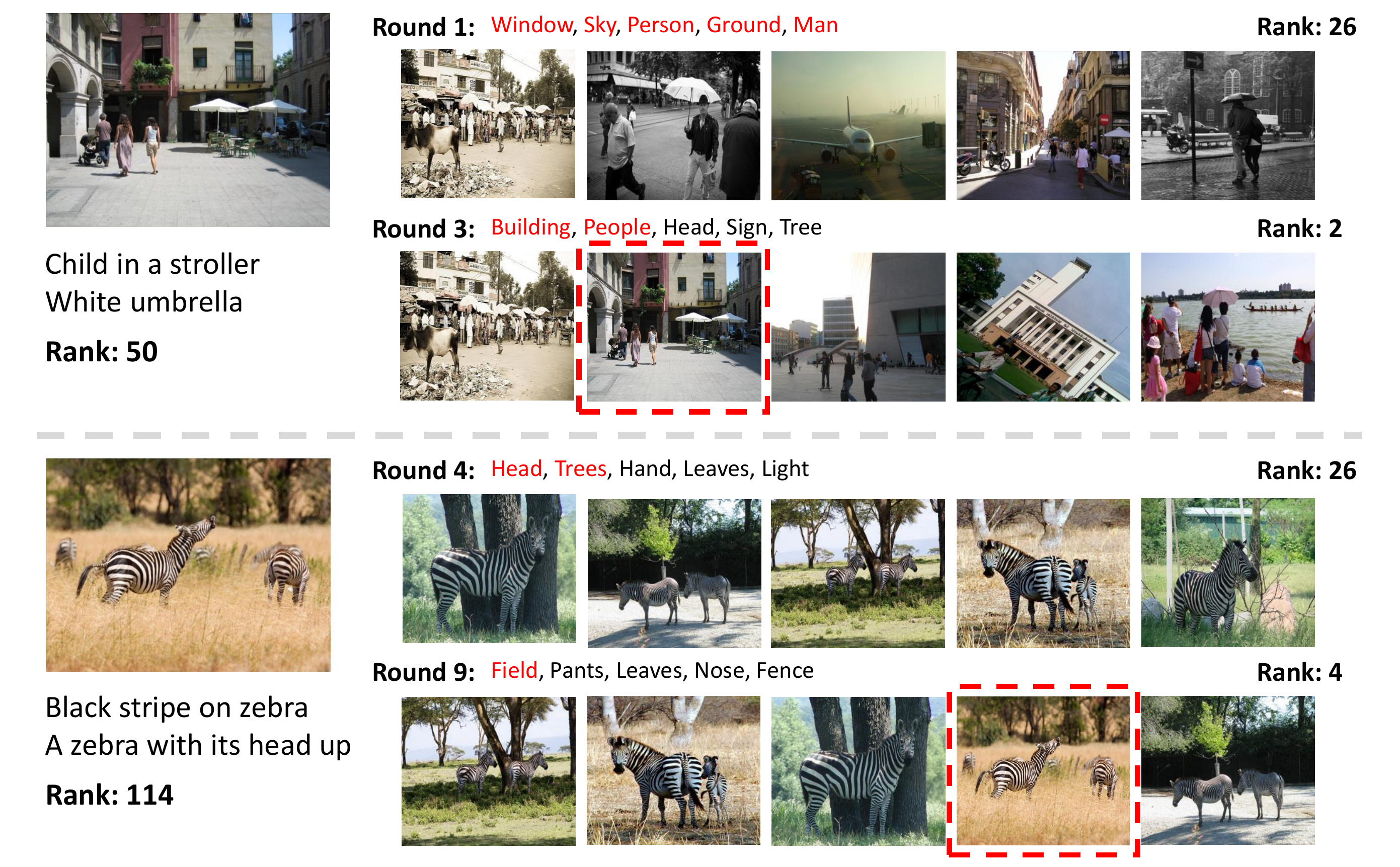}
        \vspace{-16mm}
        \caption*{}
        \end{minipage}
    }\hspace{-15mm}\vspace{-3mm}
    \subfigure[Query 4/Action 3]{
        \begin{minipage}[t]{1\linewidth}
        \centering
        \includegraphics[width=3.3in]{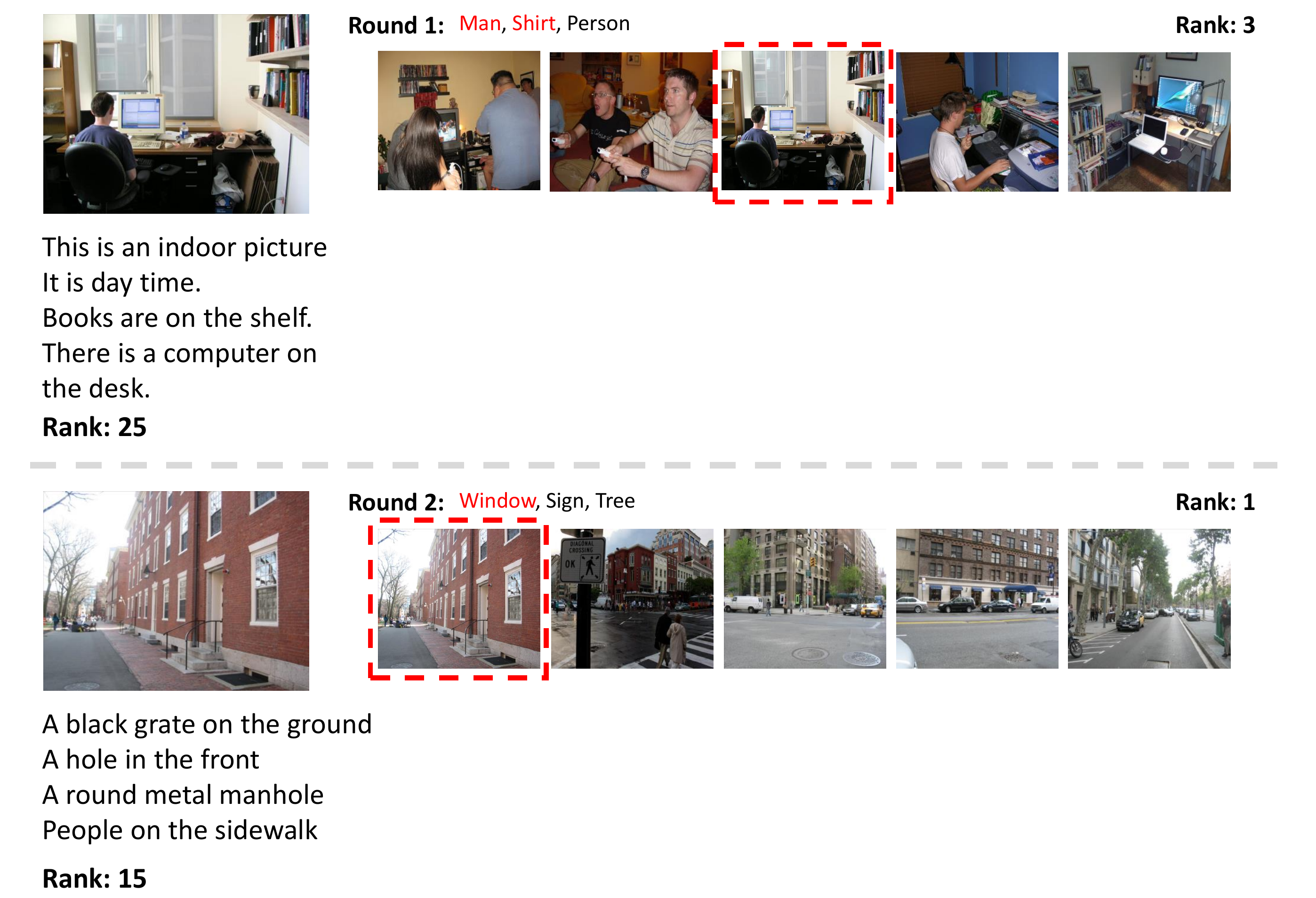}
        \vspace{-16mm}
        \caption*{}
        \end{minipage}
    }
    \vspace{-3mm}
    \caption{Visualization of Ask\&Confirm based on S-SCAN. We show examples in three settings. Positive objects in each round are highlighted in red. The target image is surrounded with a red bounding box.}
    \vspace{-7mm}
    \label{fig:vis}
\end{figure}

\subsection{Results}
\label{results}

\renewcommand\arraystretch{0.80}
\begin{table}
\begin{center}
\resizebox{\linewidth}{12mm}{
\begin{tabular}{|l|c|c|c|c|c|c|}
\hline
Method & R@1 & R@5 & R@10 & MR & Q & A \\
\hline\hline
S-SCAN& 4.5 & 13.6 & 20.4 & 416.0 & 1 & 10\\
S-SCAN+AC& 8.6 & 33.9 & 59.8 & 96.0 & 1 & 10\\
\hline
S-SCAN& 14.7 & 31.8 & 41.7 & 166.7 & 2 & 5\\
S-SCAN+AC& 16.8 & 43.3 & 67.7 & 70.7& 2 & 5\\
\hline
S-SCAN& 33.5 & 56.2 & 65.9 & 59.0 & 4 & 3\\
S-SCAN+AC& 34.1 & 61.4 & 80.1 & 37.8 & 4 & 3\\
\hline
\end{tabular}}
\end{center}
\vspace{-5mm}
\caption{Results of Ask\&Confirm on S-SCAN after 10 rounds. AC denotes the Ask\&Confirm framework.}
\vspace{-3mm}
\label{tab:ppo}
\end{table}

\renewcommand\arraystretch{0.8}
\begin{table}
\begin{center}
\resizebox{\linewidth}{12mm}{
\begin{tabular}{|l|c|c|c|c|c|c|}
\hline
Method & R@1 & R@5 & R@10 & MR & Q & A \\
\hline\hline
T-CMPL& 3.1 & 10.5 & 16.3 & 593.4 & 1 & 10\\
T-CMPL+AC& 5.2 & 20.4 & 37.0 & 313.8 & 1 & 10\\
\hline
T-CMPL& 7.3 & 19.5 & 28.3 & 283.3 & 2 & 5\\
T-CMPL+AC& 8.6 & 26.9 & 47.2 & 211.3& 2 & 5\\
\hline
T-CMPL& 14.5 & 33.5 & 44.0 & 118.2 & 4 & 3\\
T-CMPL+AC& 15.1 & 38.6 & 59.5 & 98.7& 4 & 3\\
\hline
\end{tabular}}
\end{center}
\vspace{-5mm}
\caption{Results of Ask\&Confirm on T-CMPL after 10 rounds. AC denotes the Ask\&Confirm framework.}
\vspace{-6mm}
\label{tab:ppo_global}
\end{table}

Based on S-SCAN and T-CMPL, we build two interactive retrieval models with the proposed Ask\&Confirm framework. To prove the effectiveness of Ask\&Confirm, we test them in three settings: (1) Q1/A10, (2) Q2/A5, and (3) Q4/A3. QK means K queries are given by users in the beginning, and AK means K objects are provided by an agent in each round.  All results are recorded after 10 rounds.

Results are illustrated in Table~\ref{tab:ppo} and~\ref{tab:ppo_global}. For both basic retrieval models in three different settings, Ask\&Confirm strengthens their performance in all evaluation metrics. Ask\&Confirm enhances R@10 of S-SCAN from 20.4\% to 59.8\% and strengthens R@10 of T-CMPL by 20.7\% with Q1/A10. In the other two settings, the advantage of R@10 brought from Ask\&Confirm recedes a bit but at least achieves 14.2\%. As for R@5, Ask\&Confirm based on S-SCAN achieves 61.4\% and the one based on T-CMPL achieves 38.6\% with Q4/A3. In other settings, the enhancement of Ask\&Confirm is more obvious and even achieves 11.5\% with Q2/A5 based on S-SCAN. Both basic retrieval models are improved by Ask\&Confirm of R@1 in all settings. In particular, Ask\&Confirm based on S-SCAN achieves R@1=34.1\% with Q4/A3. With Ask\&Confirm, MR of both basic retrieval models in three settings is moved up by a large margin.
These results demonstrate the effectiveness of Ask\&Confirm.

\subsection{Visualizations}

Examples of interactive retrieval are shown in Figure~\ref{fig:vis}. Several interesting discoveries are found out in visualizations. Firstly, the agent tends to offer several objects in the first few round regularly, such as ``window", ``man", ``sky", ``head", ``tree" and so on. These are the objects that come up most frequently in Visual Genome
This is a reasonable choice because it either has a large possibility to add a ground-truth object to queries or eliminates plenty of images that include these objects. Secondly, the agent can offer objects that are not common but related to the semantics of given queries and images in latter rounds. For example, to retrieve the image that includes zebras, the agent offers ``field" and ``fence" in round 9 which rarely occur but are related to zebras. To retrieve the image with a query ``White short sleeve shirt", the agent offers ``sunglasses" and ``top" in round 5 which belong to clothing just like the query, and offers ``car"  which shows in the image. We ascribe these properties to our policy learning approach. The statistic-based shaping guides the agent to give priority to the most frequent objects and the reinforcement learning promotes objects related to the semantics of images.

\subsection{Ablation Studies}

\begin{figure*}
    \vspace{-10mm}
    \subfigure[R@5]{
        \begin{minipage}[t]{0.34\linewidth}
        \centering
        \includegraphics[width=1.8in]{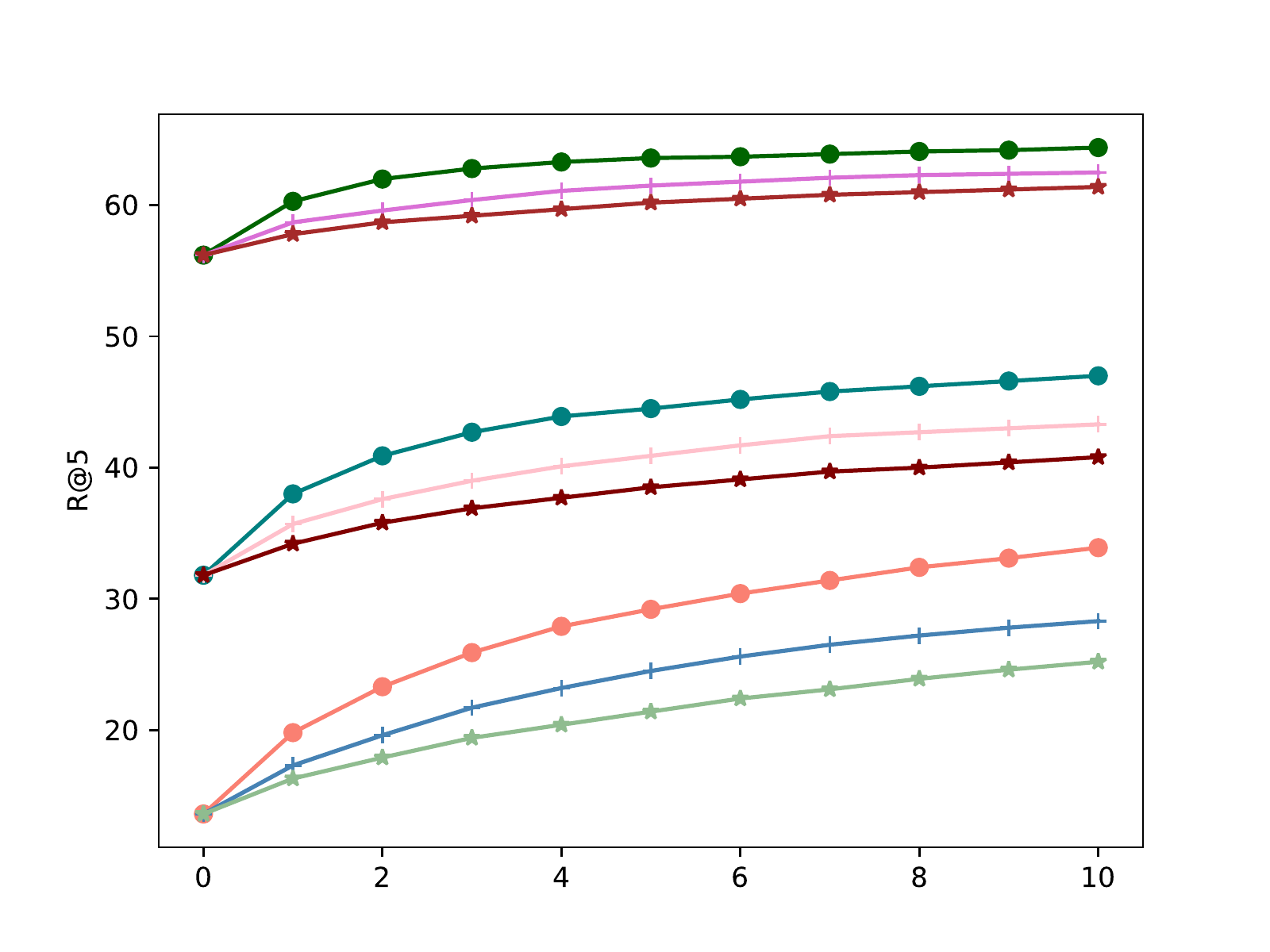}
        \vspace{-10mm}
        \caption*{}
        \end{minipage}
    }\hspace{-15mm}
    \subfigure[R@10]{
        \begin{minipage}[t]{0.34\linewidth}
        \centering
        \includegraphics[width=1.8in]{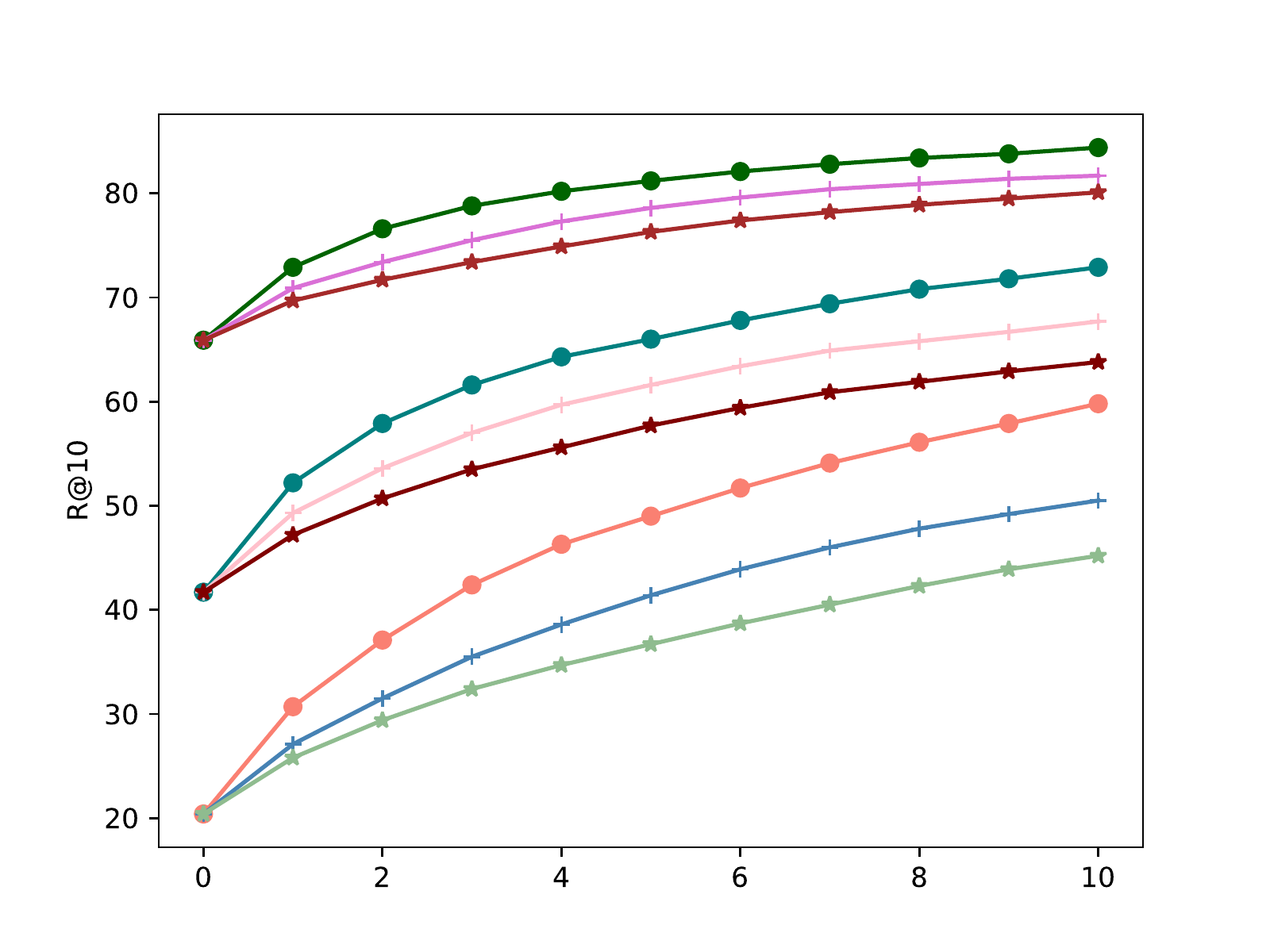}
        \vspace{-10mm}
        \caption*{}
        \end{minipage}
    }\hspace{-15mm}
    \subfigure[Mean]{
        \begin{minipage}[t]{0.34\linewidth}
        \centering
        \includegraphics[width=1.8in]{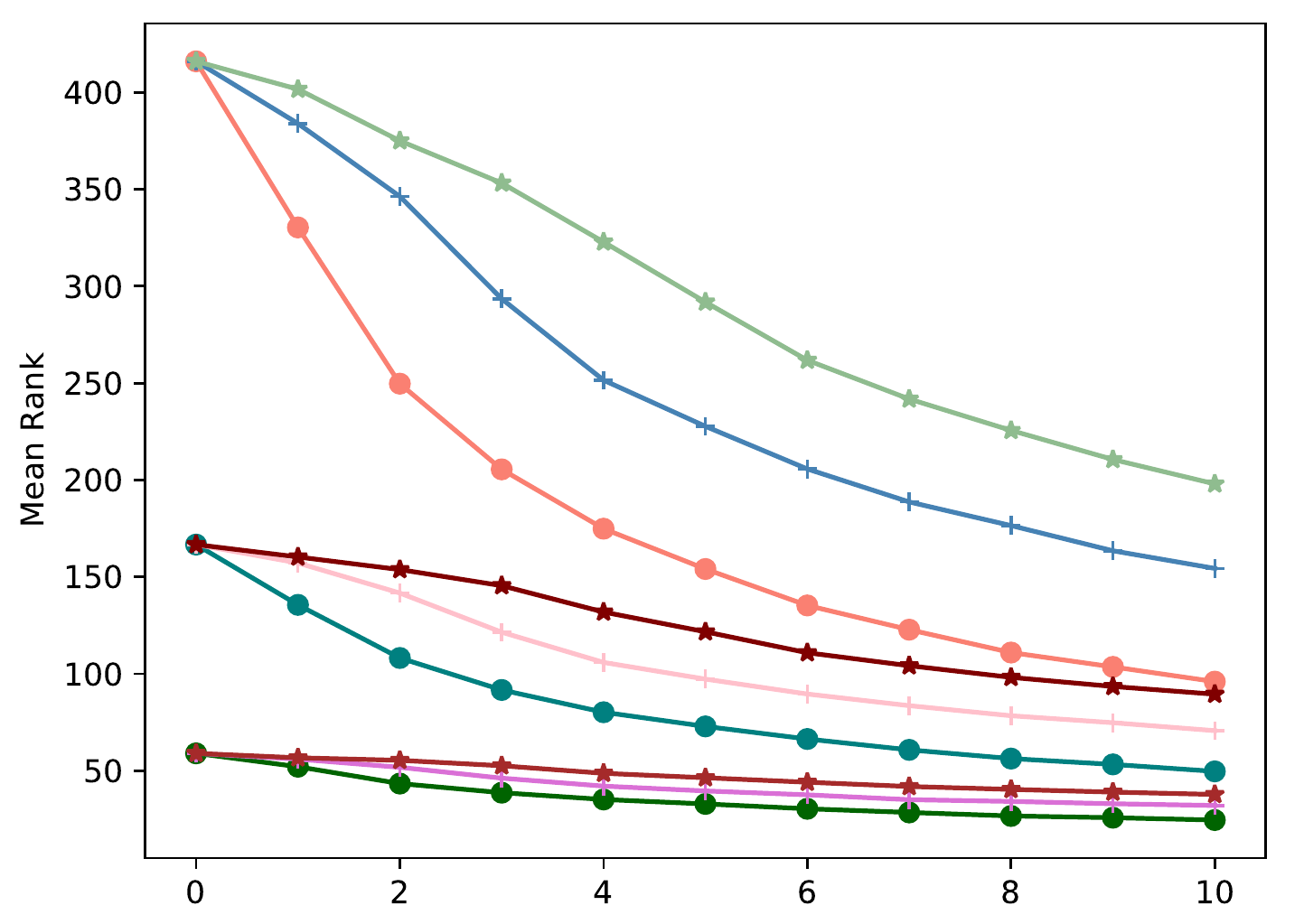}
        \vspace{-10mm}
        \caption*{}
        \end{minipage}
    }\hspace{-9mm}
    \subfigure[Setting]{
        \centering
        \begin{minipage}[t]{0.07\linewidth}
        \includegraphics[width=0.74in]{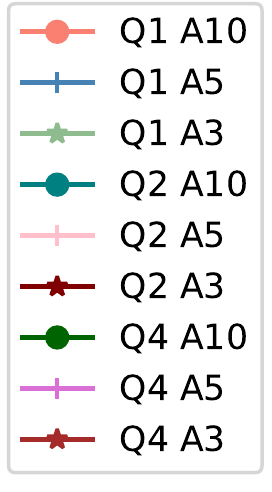}
        \vspace{-10mm}
        \caption*{}
        \end{minipage}
    }
    \vspace{-4mm}
    \caption{Results of Ask\&Confirm based on S-SCAN. The horizontal axis represents the query turn. Q denotes the number of queries and A denotes the action number in each round.}
    \vspace{-7mm}
    \label{fig:ppo}
\end{figure*}

\begin{figure*}
    \subfigure[Query 1/Action 10]{
        \begin{minipage}[t]{0.34\linewidth}
        \centering
        \includegraphics[width=1.8in]{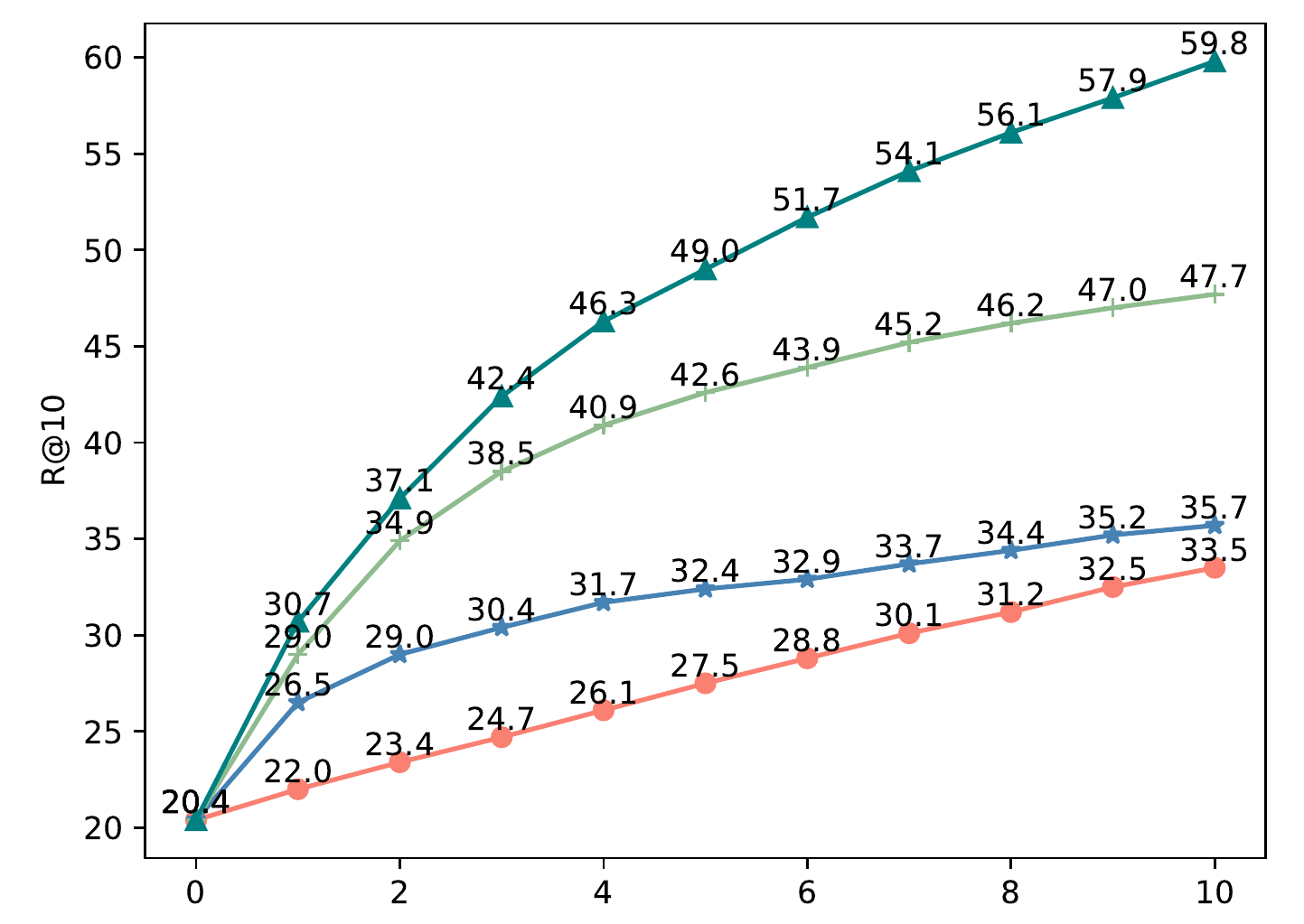}
        \vspace{-10mm}
        \caption*{}
        \end{minipage}
    }\hspace{-15mm}
    \subfigure[Query 2/Action 5]{
        \begin{minipage}[t]{0.34\linewidth}
        \centering
        \includegraphics[width=1.8in]{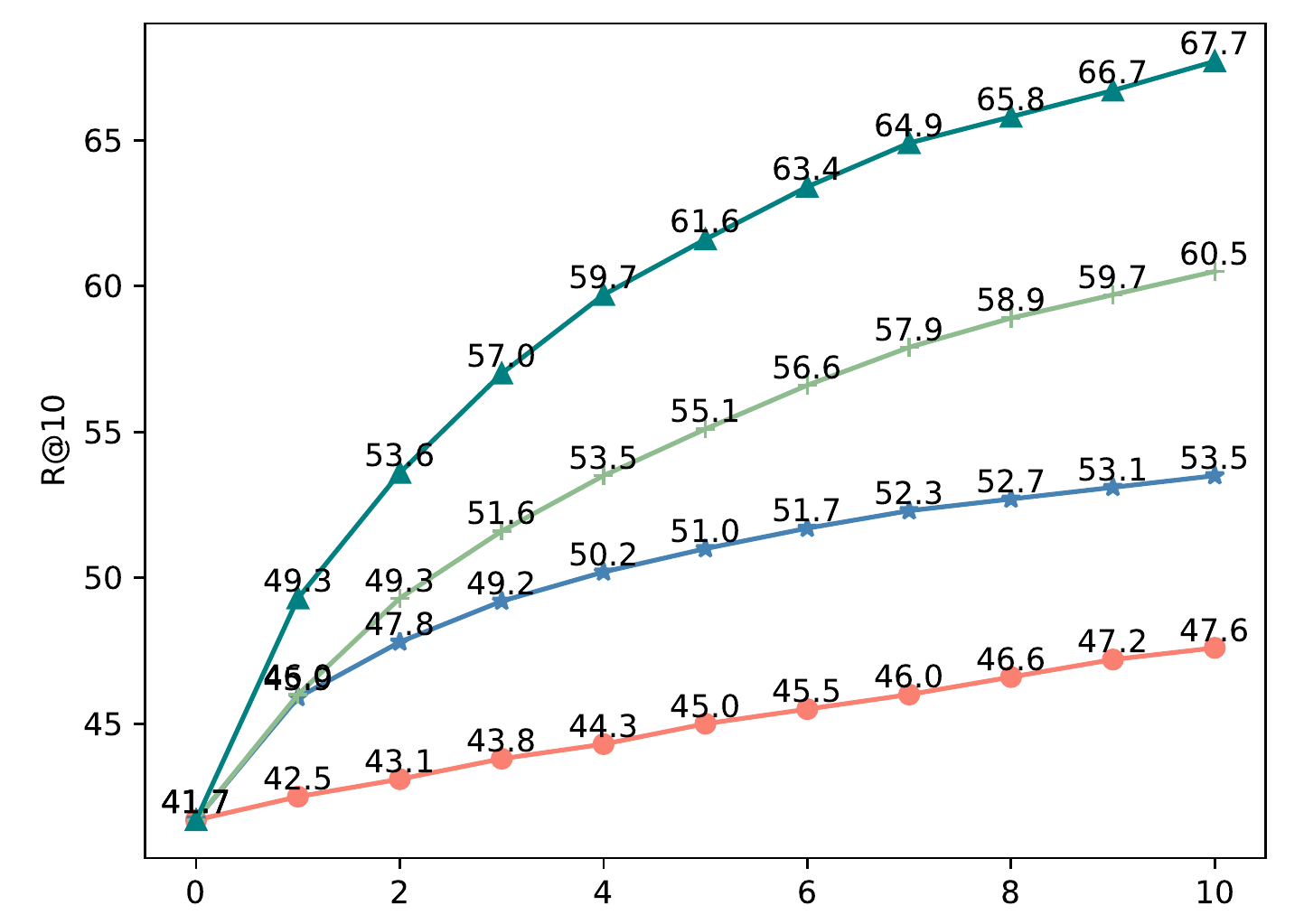}
        \vspace{-10mm}
        \caption*{}
        \end{minipage}
    }\hspace{-15mm}
    \subfigure[Query 4/Action 3]{
        \begin{minipage}[t]{0.34\linewidth}
        \centering
        \includegraphics[width=1.8in]{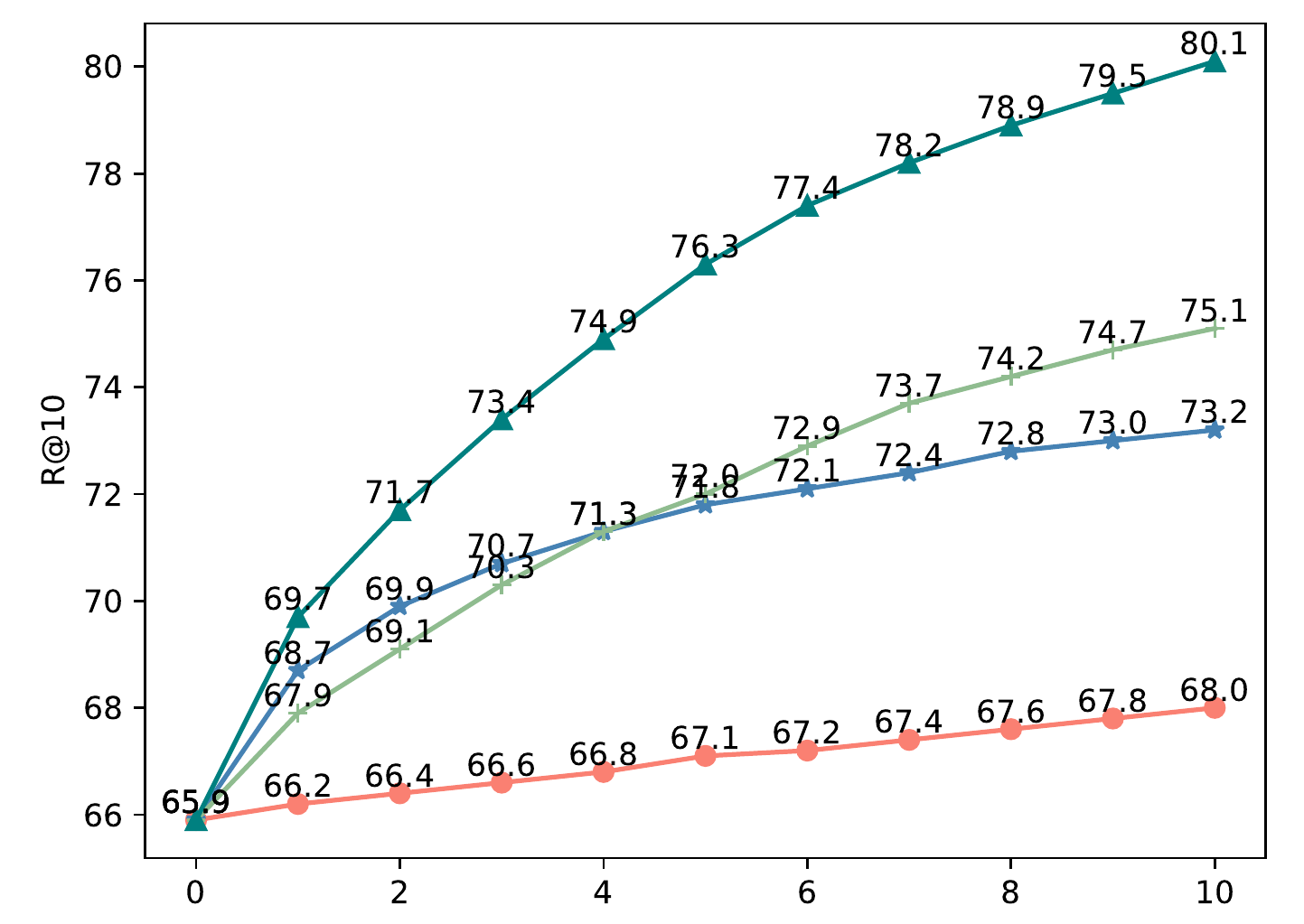}
        \vspace{-10mm}
        \caption*{}
        \end{minipage}
    }\hspace{-9mm}
    \subfigure[Policy]{
        \centering
        \begin{minipage}[t]{0.07\linewidth}
        \includegraphics[width=0.74in]{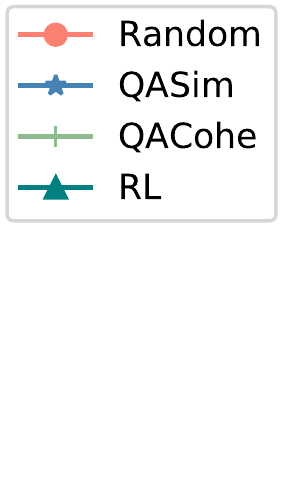}
        \vspace{-10mm}
        \caption*{}
        \end{minipage}
    }
    \vspace{-4mm}
    \caption{Results of different policies.The horizontal axis represents the query turn. The vertical axis represents R@10. The proposed RL-based policy learning approach outperforms others.}
    \label{fig:policy}
    \vspace{-4mm}
\end{figure*}

{\bf Number of Query and Action.} To verify Ask\&Confirm is robust to the number of queries and actions, we test it based on S-SCAN with different query numbers $N_Q^1\in \{1, 2, 4\}$ where users input 1, 2, or 4 queries and different action numbers $N_A\in \{3, 5, 10\}$ where the agent provides 3, 5, or 10 object candidates in each round. Results on R@5, R@10, and MR in each round are shown in Figure~\ref{fig:ppo}. 

In detail, with the same queries, the performance of Ask\&Confirm gradually improves when $N_A$ increases, which shows that more actions in each round facilitate the retrieval. 
On the other hand, when $N_A$ is fixed, queries with higher $N_Q^1=4$ outperform the ones with lower $N_Q^1$, which is consistent with our discovery in Figure~\ref{fig:partial_effect}. Although models with fewer queries achieve worse performance, improvements over them are even more. Especially, when $N_Q^1=1$ and $N_A=10$, Ask\&Confirm achieves the largest improvement. We conclude that fewer queries leave more space for the agent to optimize the basic model's retrieval. 
Despite the change of the number of queries and actions, Ask\&Confirm consistently enhances S-SCAN on all metrics. It examines the robustness of Ask\&Confirm which facilitates retrieval stably in all situations.

{\bf Policy.} 
Finding a policy that guides the agent to choose discriminative objects is essential to Ask\&Confirm. 
As a result, we compare our policy learning method  with three pre-defined policies: (1) {\bf Random}: In each round, the agent samples objects from a uniform distribution. 
(2) {\bf QASim}: Inspired by~\cite{liu2008query}, objects that have similar textual features with a query are preferred. We use cosine similarity between textual features of queries and objects to indicates their similarity. 
(3) {\bf QACohe}: Considering that some objects tend to occur coherently, such as ``building" and ``window", we compute a joint distribution $P_c(a_i,a_j)$ in the train split, where $a_i$ and $a_j$ are in the same image. Then, we use $P_c(a^{*}, a_j)$ where
$a^{*}=\argmax\limits_{a\in\mathcal{A}}\frac{1}{N_Q^t}\sum_{n=1}^{N_Q^t} cos(x^T_n, TE(\mathbb{T}(a)))$ to sample objects. 

Experiments based on S-SCAN are conducted in three settings just like Section~\ref{results}. As shown in Figure~\ref{fig:policy}, under all settings, the proposed policy learning outperforms the other by a large margin in terms of R@10. 
After 10 rounds, our policy learning strategy outperforms the second-best policy by 12.1\%, 7.2\%, and 5.0\% in three settings. We also observe that a good policy increases R@10 rapidly in the first several rounds and slows down in subsequent rounds. Such a policy provides better interactive experiences because users retrieve the target image with fewer interactions.

{\bf Model Agnostic.} By comparing the improvements on S-SCAN and T-CMPL as shown in Table~\ref{tab:imrove}, we examine that the proposed framework is model-agnostic. Although the implementation and performance of T-CMPL and S-SCAN are different, Ask\&Confirm strengthens both of them on all evaluation metrics. In detail, the two models' improvements of MR are very close. As for R@K metrics, improvements are more obvious on S-SCAN due to its better original performance. These results demonstrate that Ask\&Confirm can easily cooperate with a common text-based retrieval model to boost the retrieval performance. 

\renewcommand\arraystretch{0.8}
\begin{table}
\begin{center}
\resizebox{\linewidth}{12mm}{
\begin{tabular}{|l|c|c|c|c|c|c|}
\hline
Method & R@1 & R@5 & R@10 & MR & Q & A \\
\hline\hline
T-CMPL+AC& +1.9 & +9.9 & +10.7 & -279.6 & 1 & 10\\
S-SCAN+AC& +4.1 & +20.3 & +39.4 & -320.0 & 1 & 10\\
\hline
T-CMPL+AC& +1.3 & +7.4 & +8.9 & -72.0 & 2 & 5\\
S-SCAN+AC& +2.1 & +11.5 & +26.0 & -96.0& 2 & 5\\
\hline
T-CMPL+AC& +0.6 & +5.1 & +15.5 & -19.5 & 4 & 3\\
S-SCAN+AC& +0.6 & +15.2 & +14.2 & -21.1 & 4 & 3\\
\hline
\end{tabular}}
\end{center}
\vspace{-6mm}
\caption{Results of Ask\&Confirm on different basic retrieval models after 10 rounds.}
\label{tab:imrove}
\vspace{-6mm}
\end{table}

\subsection{User Study}
\begin{figure}
    \subfigure[User Study]{
        \begin{minipage}[t]{0.48\linewidth}
        \vspace{-9mm}
        \includegraphics[width=1.5in]{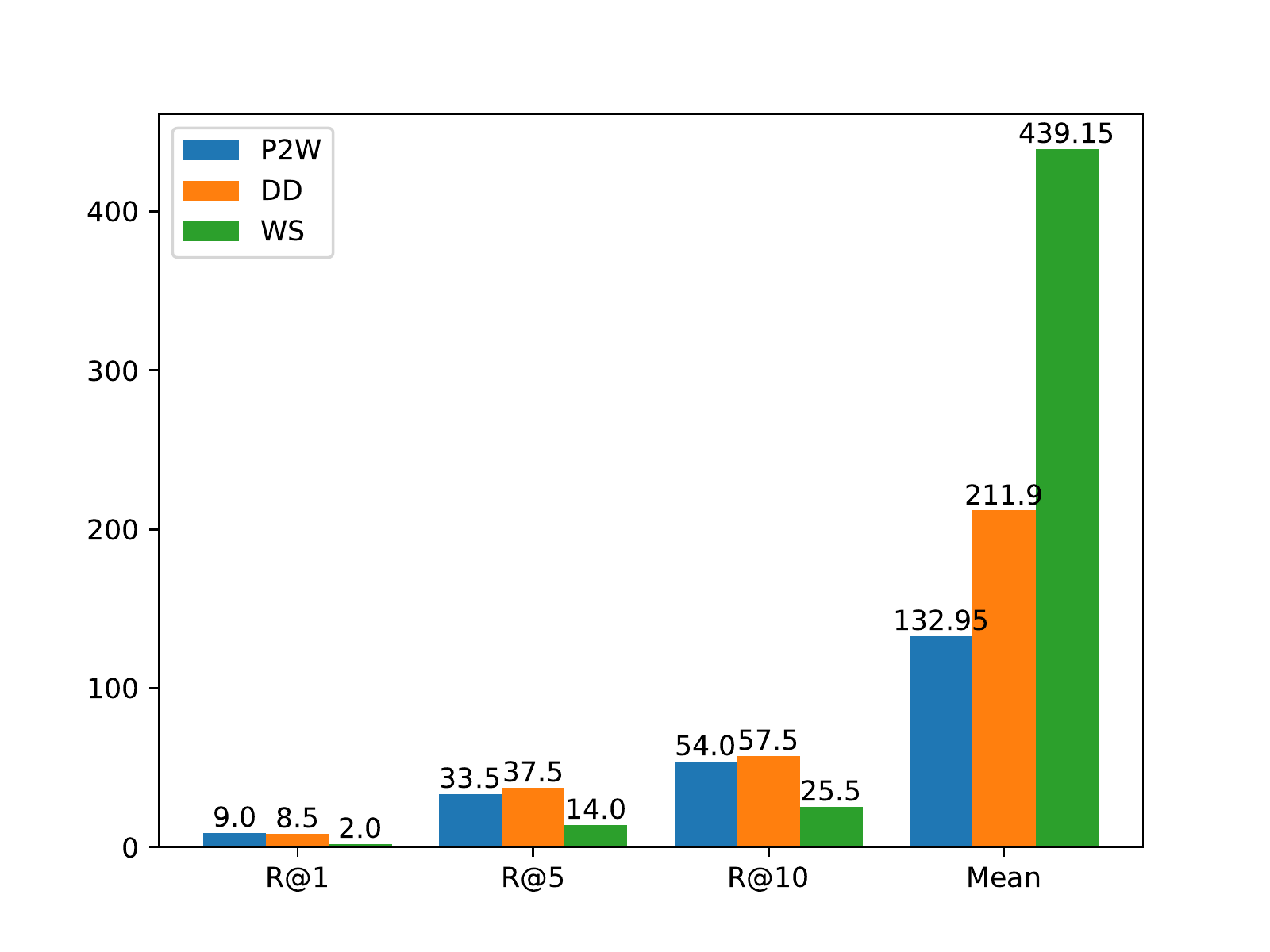}
        \vspace{-8mm}
        \caption*{}
        \end{minipage}
    }\hspace{-2mm}
    \subfigure[Object Distribution]{
        \begin{minipage}[t]{0.48\linewidth}
        \vspace{-9mm}
        \includegraphics[width=1.5in]{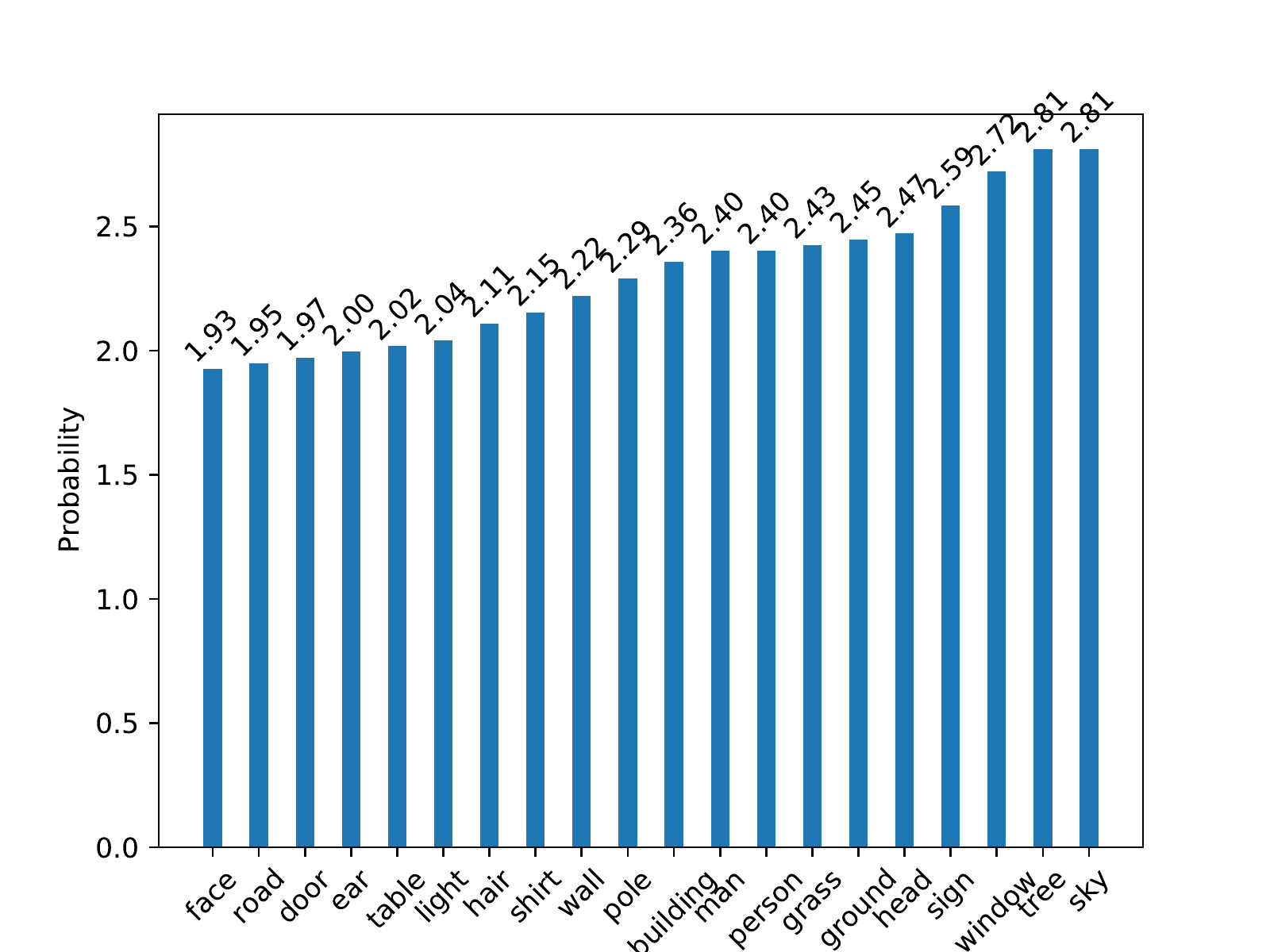}
        \vspace{-8mm}
        \caption*{}
        \end{minipage}
    }
    \vspace{-5mm}
    \caption{(a) User study. AC denotes Ask\&Confirm. (b) Object distribution during the user study.}
    \vspace{-7mm}
    \label{fig:user}
\end{figure}

To demonstrate the advantage of the active object-based interaction over tag-based and description-based interaction, we compare Ask\&Confirm (AC) with Drill-Down (DD)~\cite{tan2019drill} and WhittleSearch (WS)~\cite{kovashka2012whittlesearch} where DD is a description-based method and WS is a tag-based method. To make a fair comparison of interactive mode, we re-implement DD and WS based on S-SCAN and adopt their interactive mode. 50 images are sampled from the test set. For each image, 4 different users (details in supplementary) are required to retrieve it in 5 rounds with 3 different methods.
The retrieval performance in terms of R@1, R@5, R@10 and Mean Rank (Mean) are shown in Figure~\ref{fig:user} (a). 


To evaluate users' effort on different methods, we record the average time users take to retrieve each image. AC costs {\bf 37.67s}, DD costs {\bf 53.60s} and WS costs {\bf 35.18s}. 

\textbf{Conclusion on Performance:} Ask\&Confirm achieves similar R@k accuracy and much better Mean Rank compared to DD. Ask\&Confirm significantly outperforms WS.

\textbf{Conclusion on User Effort:} Ask\&Confirm takes significantly less time to complete the retrieval compared to DD and takes similar time compared to WS. 

Overall, Ask\&Confirm achieves similar performance with description-based interaction and similar retrieval time with tag-based interaction. It examines that Ask\&Confirm not only achieves a friendly user experience, but also achieves excellent retrieval performance. Furthermore, Figure~\ref{fig:user} (b) shows the percentage of objects provided by Ask\&Confirm in the user study. It demonstrates what the RL-based policy learns from the image gallery. 

\section{Conclusion}
We firstly introduce the partial-query problem that easily makes cross-modal retrieval models collapsed and propose Ask\&Confirm, an interactive retrieval framework, to tackle this problem. Ask\&Confirm heuristically guides users to enrich details of images by actively searching for discriminative objects of the target image for users to confirm. A weakly-supervised RL-based policy is proposed to conduct the active search, which leverages the characteristics of the image gallery.
Experimental results demonstrate the effectiveness and robustness of Ask\&Confirm. The weakly-supervised training procedure also makes it more practical than other dialog-based retrieval models.  

\section{Acknowledgement}
This work is supported by Joint Funds of the National Science Foundation of China under Grant U18092006, the Shanghai Municipal Science and Technology Committee of Shanghai Outstanding Academic Leaders Plan under Grant 19XD1434000, the Projects of International Cooperation of Shanghai Municipal Science and Technology Committee under Grant 19490712800, the National Natural Science Foundation of China under Grant 61772369, Grant 61773166, Grant 61771144, National Key R\&D Program of China under Grant 2020YFA0711400, Shanghai Municipal Science and Technology Major Project (2021SHZDZX0100), Shanghai Municipal Commission of Science and Technology Project (19511132101), the Changjiang Scholars Program of China, and the Fundamental Research Funds for the Central Universities.

{\small
\bibliographystyle{ieee_fullname}
\bibliography{egbib}
}

\newpage
\section{Supplementary}
\subsection{Network Implementation Details}

In this section, we describe implementation details of the parameterized components in Ask\&Confirm: Text Encoder, Image Encoder, policy net, and value net. 

{\bf Text Encoder.} We map the natural language to a $256$-dimensional vector space. Given a sentence $T$ that contains $n$ words, we represent the $i$ th word in it with a one-hot vector showing the index of the word in a vocabulary and then embed the word into a $300$-dimensional vector $x_i$ through an embedding matrix $W_e$. Then, we use a one-layer unidirectional GRU to map the vector to the final textual feature along with the sentence context. The GRU reads the sentence $T$ from $1$ to $n$ th word and obtains the final textual feature $x^T$:
\begin{equation}
    x^T = GRU(x_i), i\in [1,n]
\end{equation}

We do not use a bidirectional GRU in this work because the performance between them is close according to our experimental results.

{\bf Image Encoder.} Given an image $I$, we aim to map it to a set of $256$-dimensional vectors $X^I=\{x^I_1,x^I_2, ..., x^I_k\}, k=36$ where each vector encode a region and predict a set of objects $A=\{a_1, a_2, ..., a_j\}$ in an image. We refer to detection of salient regions as bottom-up attention~\cite{anderson2018bottom} and implement it with a Faster-RCNN~\cite{ren2015faster}. We adopt the Faster-RCNN whose backbone is a ResNet-101~\cite{he2016deep} pretrained by Anderson et al.~\cite{anderson2018bottom} on Visual Genome~\cite{krishna2017visual}. For each region $i$, $f_i$ is defined as the mean-pooled feature from this region and the dimension of $f_i$ is 2048. To get a $256$-dimensional vector as textual vectors, we add an two-layer MLP to transform $f_i$ to $x^I_i$:
\begin{equation}
    x^I_i=MLP(f_i)
\end{equation}

As for predicting objects, the original model predicts attribute classes and instance classes together to learn feature representations with rich semantic meaning. However, in our Ask\&Confirm, we just need objects in an image. Hence, we re-train a two-layer MLP to predict the objects in $I$. After obtaining $X^I$, we concatenate all vectors into a $36\times 256$-dimensional vector $X^I_1$ and use the MLP to predict every object's probability of being in $I$. The architecture of the re-trained MLP is shown in Table~\ref{tab:ie}. 

\begin{table}
\begin{center}
\begin{tabular}{|l|c|c|}
\hline
Type & Weight shape & Input size \\
\hline\hline
Fc & 2048$\times$256 & N$\times$2048 \\
\hline
Fc & 256$\times$256 & N$\times$256 \\
\hline  
Fc & 9216$\times$ 256 & N$\times$9216\\
\hline
Fc & 256$\times$ 1601 & N$\times$256 \\
\hline
\end{tabular}
\end{center}
\caption{The architecture of MLP that predicts the objects in an image. N denotes the batchsize and Fc denotes the fully-connected layer.}
\label{tab:ie}
\end{table}

\begin{table}
\begin{center}
\begin{tabular}{|l|c|c|}
\hline
Type & Weight shape & Input size \\
\hline\hline
Fc & 3202$\times$ 256 & N$\times$3202\\
\hline
Tanh & - &N$\times$256 \\
\hline
Fc & 256$\times$256 & N$\times$256 \\
\hline
Tanh & - &N$\times$256 \\
\hline
Fc & 256$\times$ 1601 & N$\times$256 \\
\hline
Softmax & -& N$\times$1601 \\
\hline
\end{tabular}
\end{center}
\caption{The architecture of policy net. Tanh denotes the hyperbolic tangent function and Softmax denotes the softmax function.}
\label{tab:p}
\end{table}

\begin{table}
\begin{center}
\begin{tabular}{|l|c|c|}
\hline
Type & Weight shape & Input size \\
\hline\hline
Fc & 3202$\times$ 256 & N$\times$3202\\
\hline
Tanh & - &N$\times$256 \\
\hline
Fc & 256$\times$1 & N$\times$256 \\
\hline
\end{tabular}
\end{center}
\caption{The architecture of value net.}
\label{tab:v}
\end{table}

{\bf Policy Net.} Given a state $s\in\mathbbm{R}^{3202}$. The policy net $\pi$ outputs a $1601$-dimensional vector as the object sample distribution. During training, we apply a stochastic sampling to choose objects to users while in the testing period, a greedy sampling is applied. The architecture of $\pi$ is shown in Table~\ref{tab:p}. 

{\bf Value Net.} Given a state $s\in\mathbbm{R}^{3202}$. The value net $V$ outputs a scalar that estimates the real advantage returned by the interactive agent. According to~\cite{schulman2017proximal}, estimating the advantage is helpful to reduce the variance of reinforcement learning. The architecture of $V$ is shown in Table~\ref{tab:v}. 

\subsection{Implementation Details of Partial Query v.s. Partial Query + Objects}

To demonstrate that objects in an image are discriminative enough to distinguish different images, we conduct an experiment, i.e. partial query v.s. partial query + objects, to compare two types of queries: partial query and supplement partial query with the name of the objects. The experiment is evaluated on Visual Genome~\cite{krishna2017visual}.

In detail, for each image $i$ that includes several captions $Q=\{q_n\}_{n=1}^{N_Q}$ to describe it, we randomly choose one caption $q_n$ as the partial query. As for the additional objects, we use an object detector~\cite{anderson2018bottom} pretrained on Visual Genome~\cite{krishna2017visual} to detect all objects $A=\{a_n\}_{n=1}^{N_A}$ contained in each image. These objects' names are regarded as additional queries. For example, if an initial query $q$, i.e. ``a man is surfing", is chosen to retrieve its corresponding image $i$ and the detector detects all objects $A$, i.e. ``man", ``sea" and ``surfboard", in the target image, these words of objects are regarded as three individual queries appended to the initial query. Thus, the new queries includes four query: ``a man is surfing", ``man", ``sea" and ``surfboard". As a result, the new query adds more discriminative information to retrieve the target image. 

\subsection{Pseudo Code}

To describe our Ask\&Confirm in more detail, we give the pseudo code of the whole workflow of Ask\&Confirm as shown in Algorithm~\ref{alg:1}.

\begin{algorithm}[H]
\caption{The whole workflow of Ask\&Confirm}
\label{alg:1}
\begin{algorithmic}
\State Initialize Text Encoder $TE$ and Image Encoder $IE$
\State Initialize policy parameters $\phi_\pi$ and value parameters $\phi_V$
\State\textbf{Input:} $I=\{i_n\}_{n=1}^{N}$: the whole gallery images
\For{episode=1,M}
\State\textbf{Input:} $i_*$: the target image 
\State\textbf{Input:} $Q_1=\{q_n\}^{N^1_Q}_{n=1}$: a set of input partial queries
\For{$t$=1, $T$} 
\For{$n$=1, $N^t_Q$}
\State $x_n^T=TE(q_n)$
\EndFor
\State Obtain textual features $X_t^T=\{x_n^T\}_{n=1}^{N_Q^T}$
\For{$n$=1, $N$}
\State $(x^I_n, A_n)=IE(i_n)$
\State Compute similarity $S_{t,n}(X^T_t, x^I_n)$
\EndFor
\State\textbf{Question:} Object candidates: $A_t=\{a_n\}_{n=1}^{N_A}$
\State\textbf{Feedback:} Positive objects: $A_t^p=\{a_n^p\}^{N_A^p}_{n=1}$
\State\textbf{Feedback:} Negative objects: $A_t^q=\{a_n^q\}^{N_A^q}_{n=1}$
\For{$n$=1, $N$}
\State Refine $S_{t,n}=S_{t,n}\times 0.9,if\;A_n\cap A_t^q\neq\emptyset$
\EndFor
\State\textbf{Output:} $i_t=\argmax\limits_{i_n}S_{t,n}$
\State Update queries $Q_{t+1}=Q_{t}\cup A_t^p$
\EndFor
\If{episode $\%N_s==0$}
\State Collect a set of episode
\State Run PPO to optimize $\phi_\pi$ and $\phi_V$
\EndIf
\EndFor
\end{algorithmic}
\end{algorithm}

\subsection{Details of each user}

\begin{figure}[h]
    \subfigure[Mean Rank]{
        \begin{minipage}[t]{\linewidth}
        \centering
        \includegraphics[width=3in]{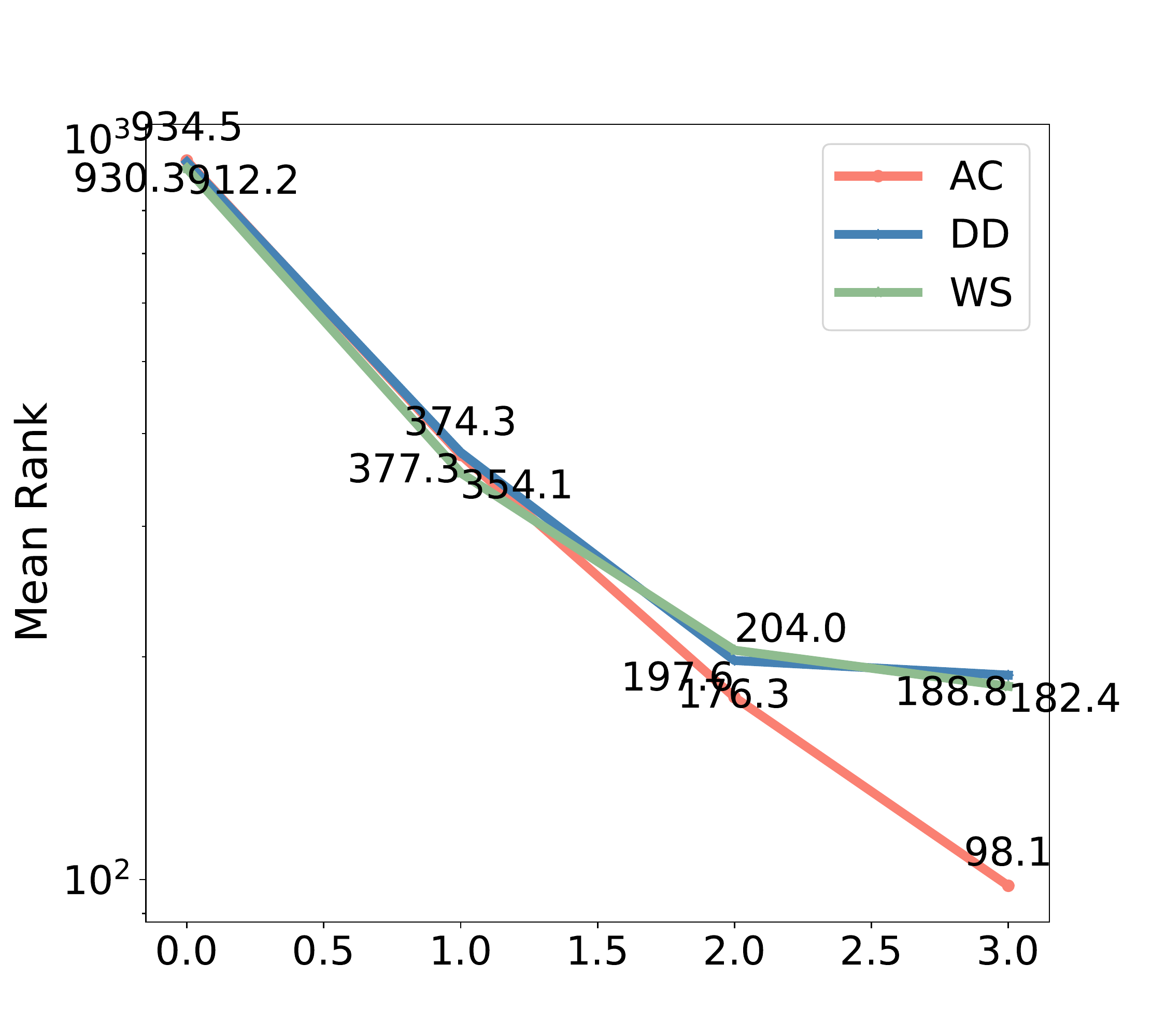}
        \caption*{}
        \end{minipage}
    }
    \subfigure[R@10]{
        \begin{minipage}[t]{\linewidth}
        \centering
        \includegraphics[width=3.0in]{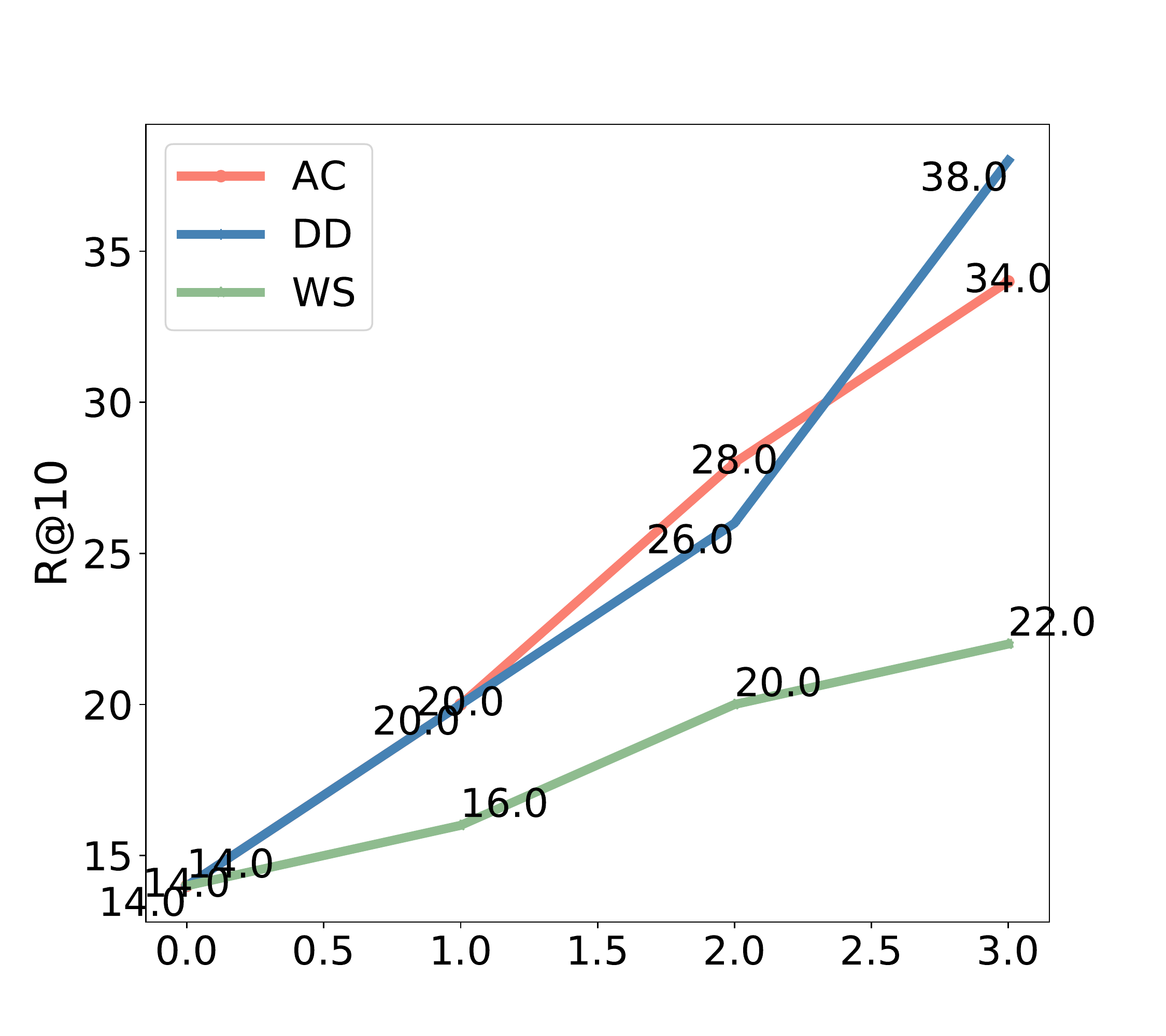}
        \caption*{}
        \end{minipage}
    }
    \caption{Mean Rank (lower is better) and R@10 (higher is better) over iterations of different interactive image retrieval systems. Ac denotes Ask\&Confirm, DD denotes Drill Down and WS denotes WhittleSearch.}
    \label{fig:iteration}
\end{figure}

In this section, we give an detailed description of the user study. For the user selection, we follow the metric in Drill Down. An expert user (male) familiar with interactive retrieval, and three novice users (1 female+2 male) are selected. They are volunteering postgraduates. For fair comparison, users are blind to these methods. After showing the target image for 5 sec, a user is asked to retrieve by interacting with the retrieval system. For AC, top-10 retrieved images and object candidates are shown to the user per turn as hints, and the user confirms the presence of objects. Detailed results are shown in Table.\ref{tab:user_performance} where Exp denotes the expert user and Nov denotes the novice user. 

Furthermore, we conduct the evaluation over iterations of different interactive image retrieval systems to compare their performance. Results are shown in Figure.\ref{fig:iteration}. AC obtains similar performance over iterations compared with DD and it outperforms WS with a large margin. As for Mean Rank, AC outperforms other approaches. Considering that AC costs much less time than DD, we conclude that AC performs the best among the three approaches.

Meanwhile, to give a subjective comparison of user experience, we conduct a post-experiment survey to acquire users' feeling about different interactive retrieval methods. Three of the users (1 Exp+2 Nov) prefer AC and a novice user prefers DD. 
The reason that they think WS is not user-friendly is the poor performance such that they can hardly find the target image.

\subsection{AC v.s. QACohe}

\begin{figure*}
    \vspace{-6mm}
    \subfigure[Query 1/Action 10]{
        \begin{minipage}[t]{0.39\linewidth}
        \centering
        \includegraphics[width=2.2in]{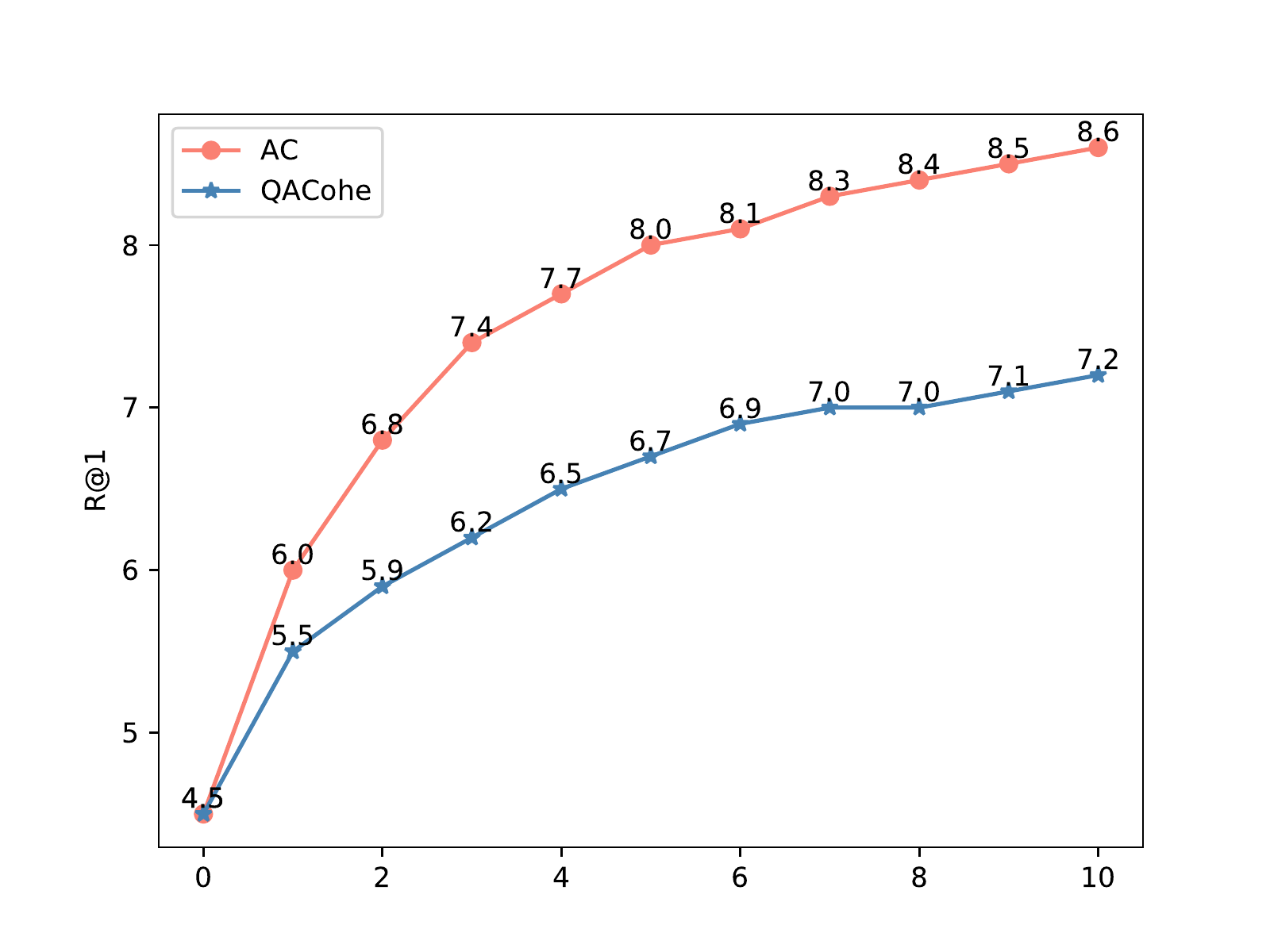}
        \vspace{-10mm}
        \caption*{}
        \end{minipage}
    }\hspace{-15mm}
    \subfigure[Query 2/Action 5]{
        \begin{minipage}[t]{0.39\linewidth}
        \centering
        \includegraphics[width=2.2in]{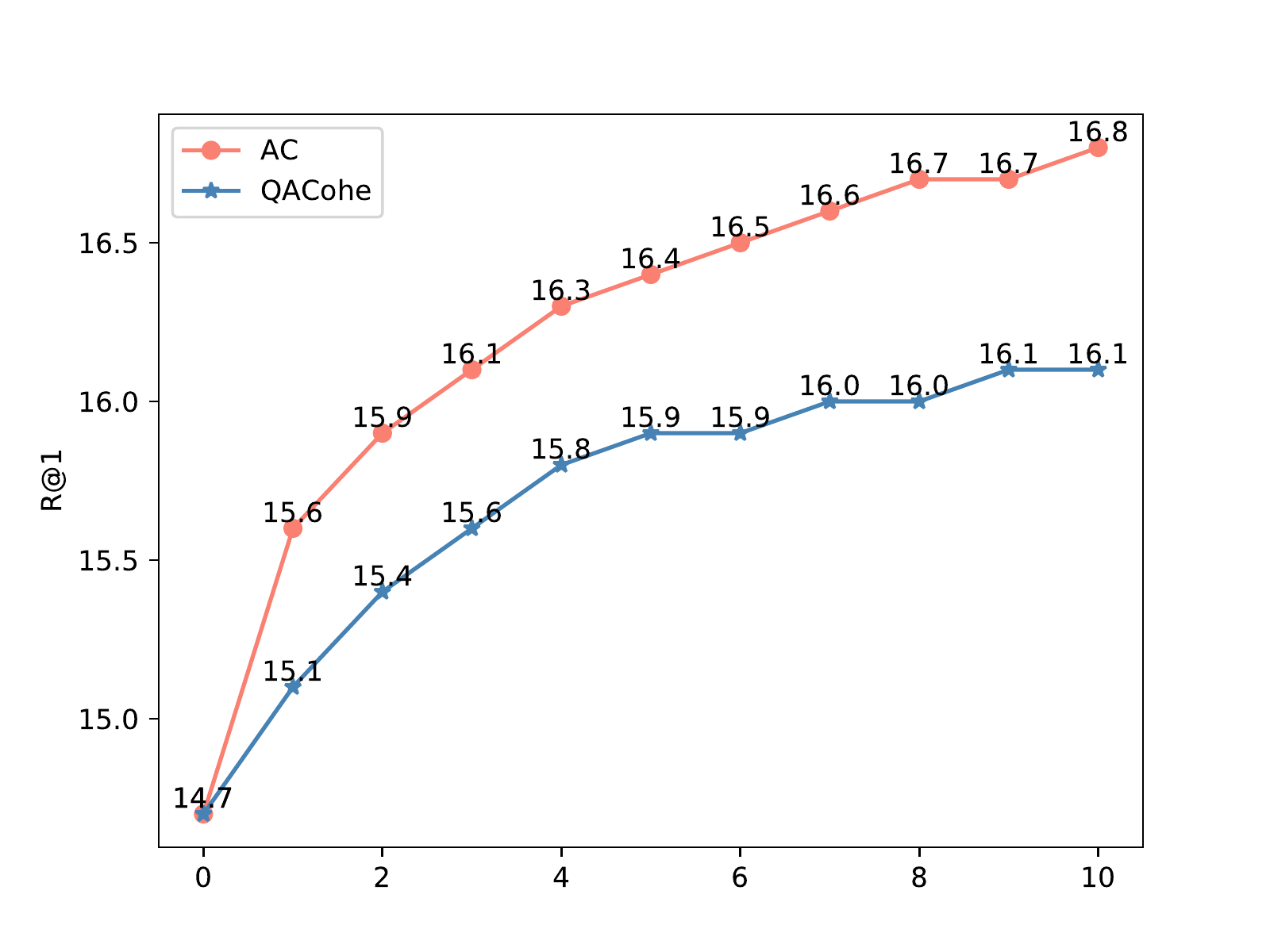}
        \vspace{-10mm}
        \caption*{}
        \end{minipage}
    }\hspace{-15mm}
    \subfigure[Query 4/Action 3]{
        \begin{minipage}[t]{0.39\linewidth}
        \centering
        \includegraphics[width=2.2in]{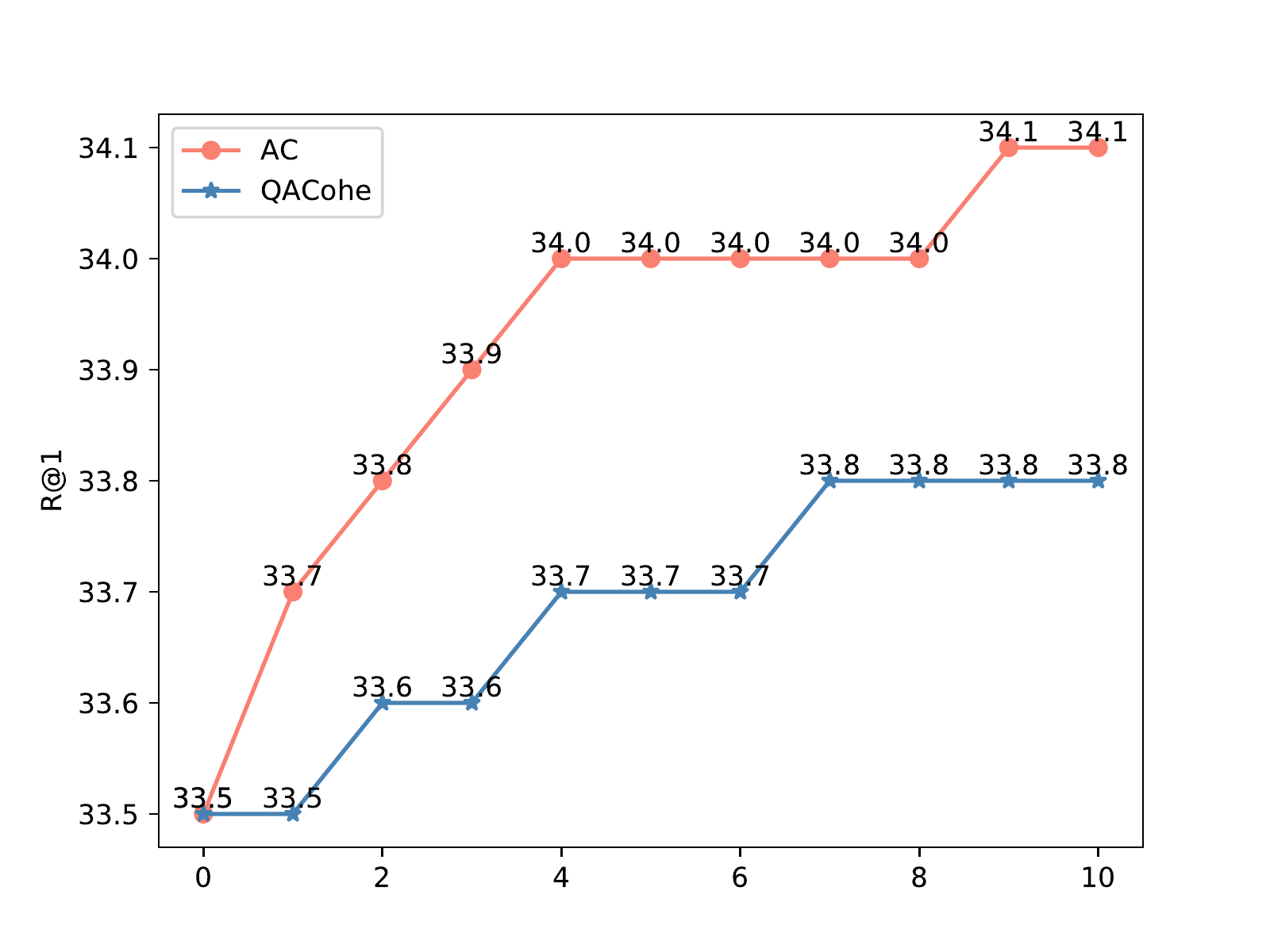}
        \vspace{-10mm}
        \caption*{}
        \end{minipage}
    }\hspace{-9mm}
    \vspace{-4mm}
    \caption{Results of AC and QACohe on R@1.The horizontal axis represents the query turn.}
    \vspace{-7mm}
    \label{fig:ac_qacohe_r1}
\end{figure*}

\begin{figure*}
    \subfigure[Query 1/Action 10]{
        \begin{minipage}[t]{0.39\linewidth}
        \centering
        \includegraphics[width=2.2in]{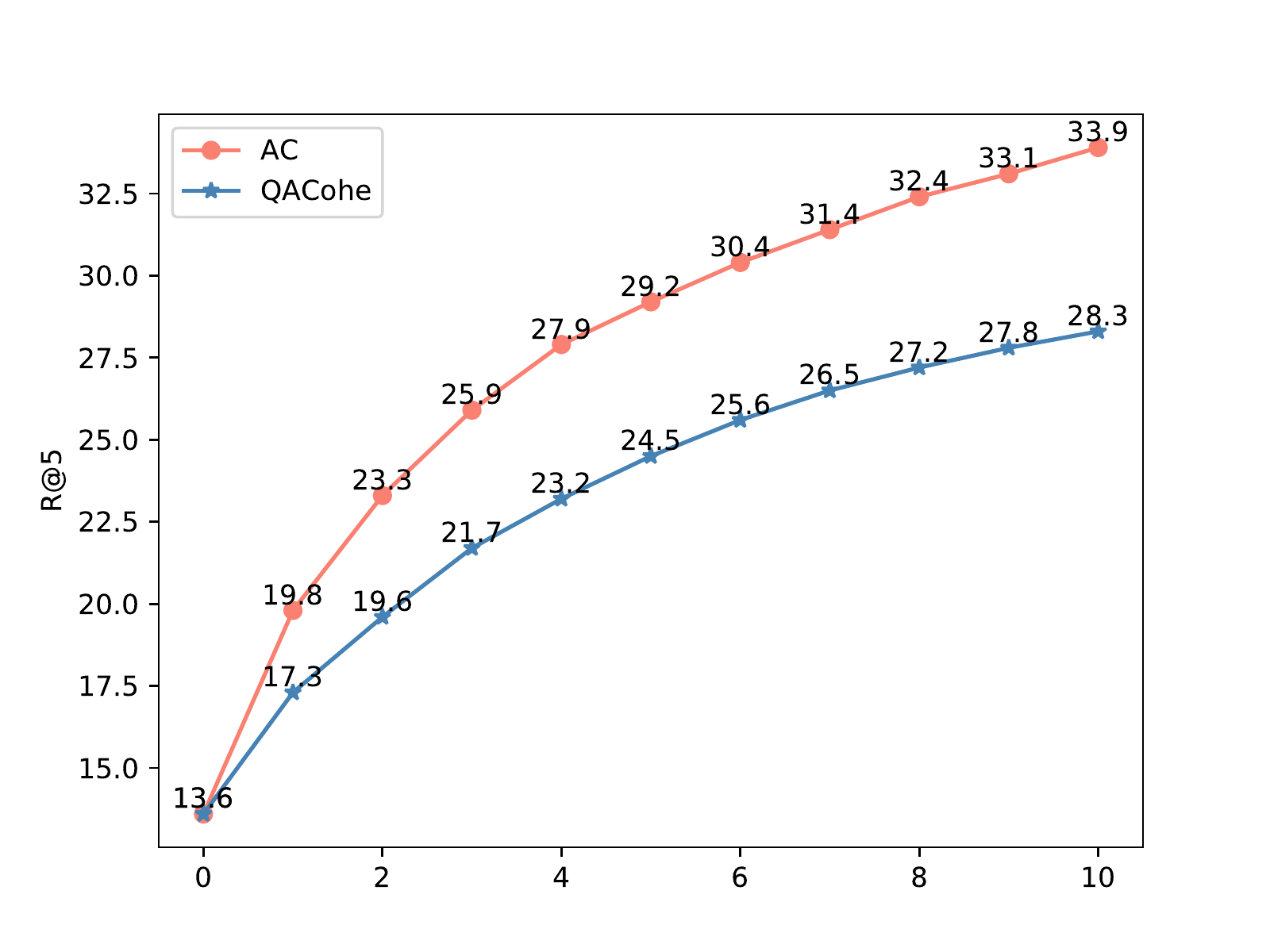}
        \vspace{-10mm}
        \caption*{}
        \end{minipage}
    }\hspace{-15mm}
    \subfigure[Query 2/Action 5]{
        \begin{minipage}[t]{0.39\linewidth}
        \centering
        \includegraphics[width=2.2in]{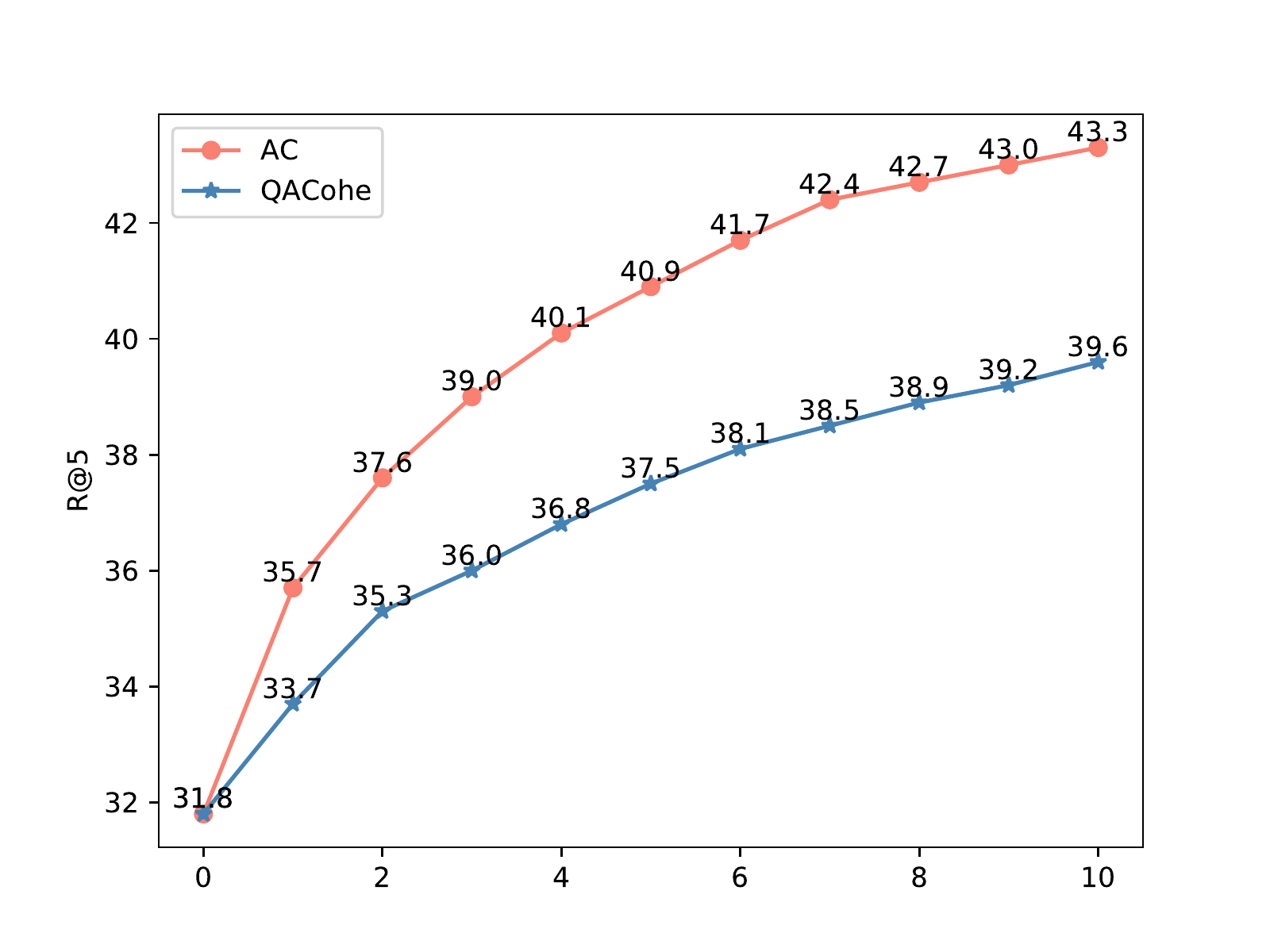}
        \vspace{-10mm}
        \caption*{}
        \end{minipage}
    }\hspace{-15mm}
    \subfigure[Query 4/Action 3]{
        \begin{minipage}[t]{0.39\linewidth}
        \centering
        \includegraphics[width=2.2in]{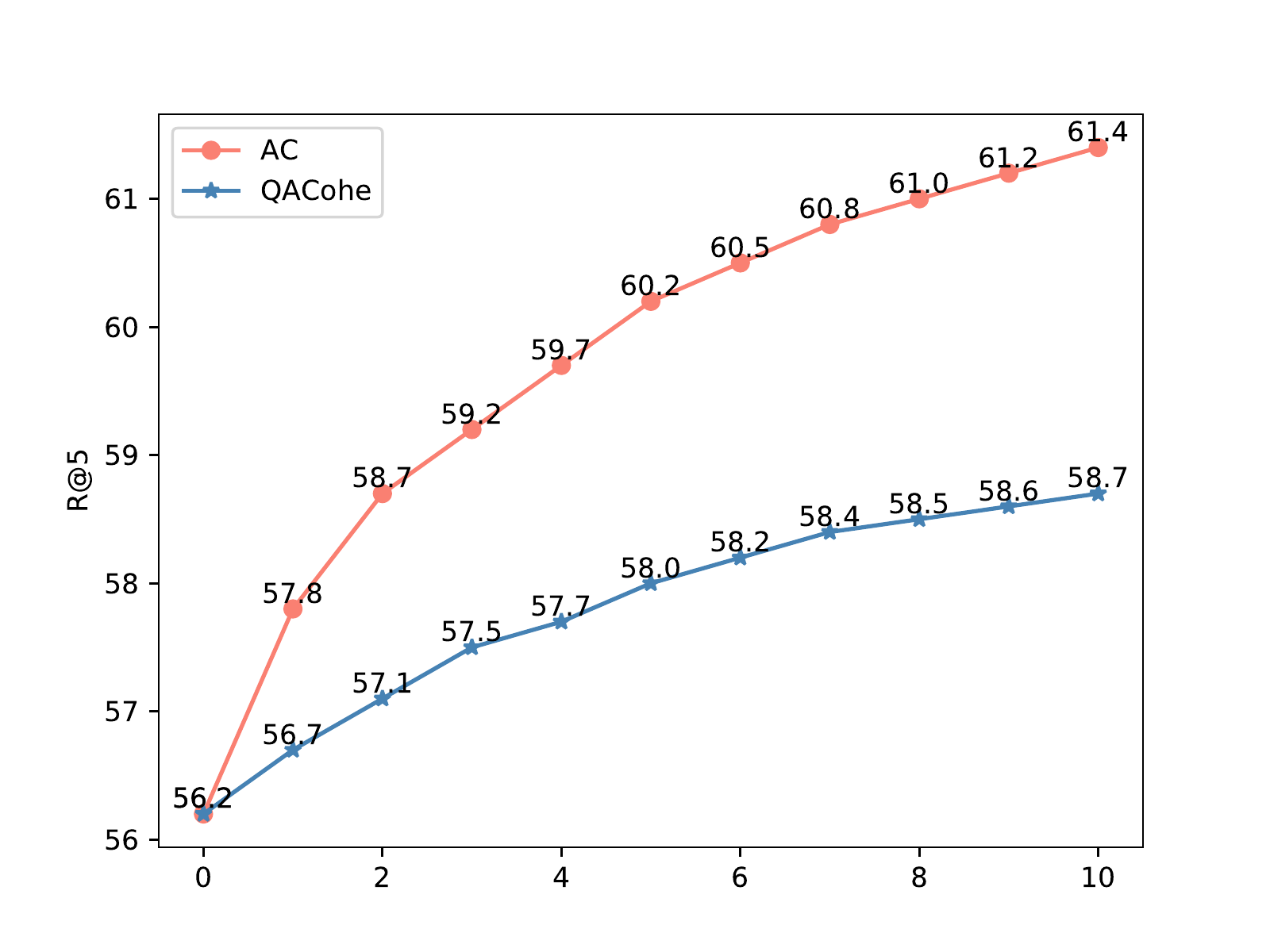}
        \vspace{-10mm}
        \caption*{}
        \end{minipage}
    }\hspace{-9mm}
    \vspace{-4mm}
    \caption{Results of AC and QACohe on R@5.The horizontal axis represents the query turn.}
    \vspace{-7mm}
    \label{fig:ac_qacohe_r5}
\end{figure*}

\begin{figure*}
    \subfigure[Query 1/Action 10]{
        \begin{minipage}[t]{0.39\linewidth}
        \centering
        \includegraphics[width=2.2in]{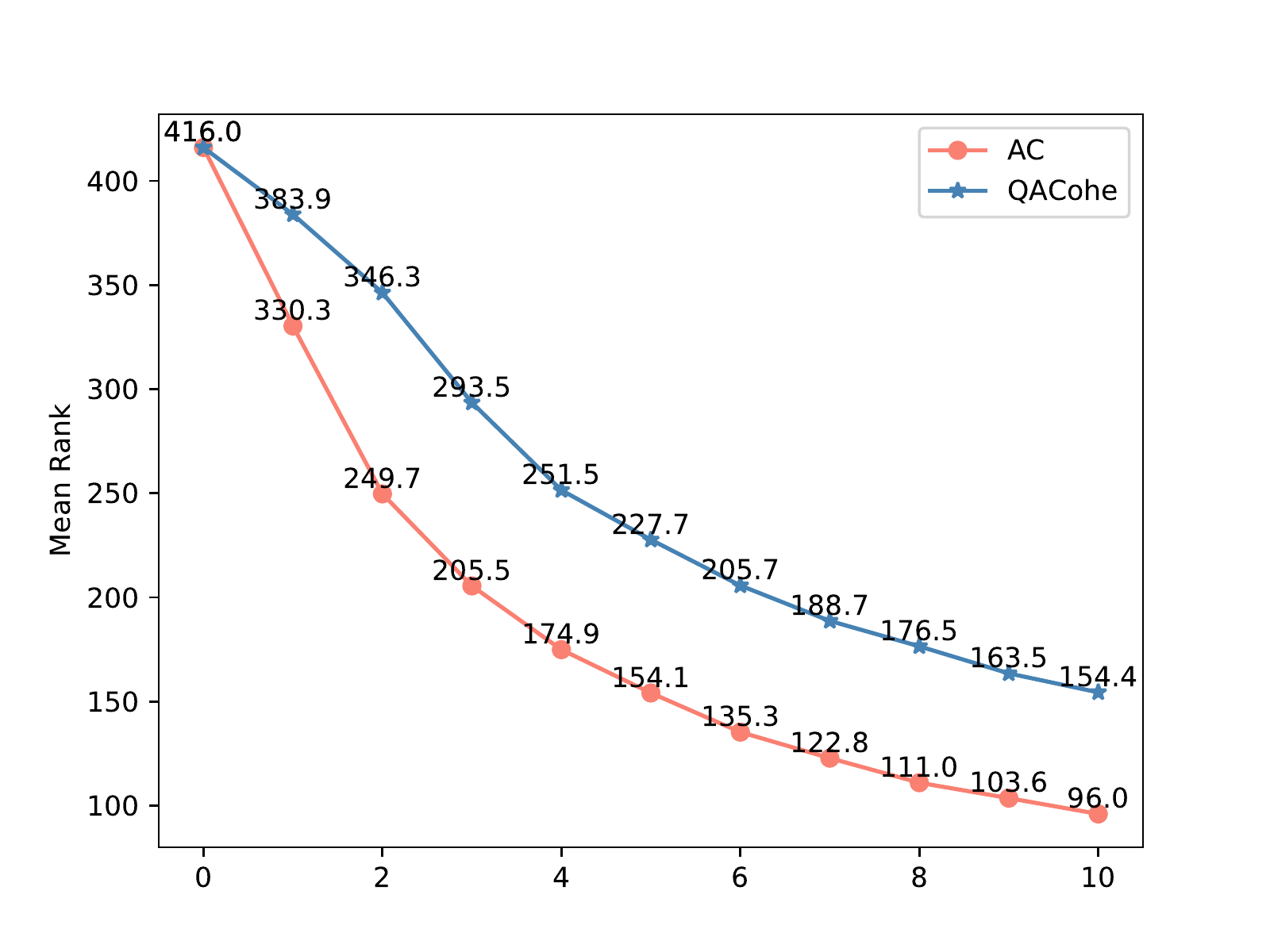}
        \vspace{-10mm}
        \caption*{}
        \end{minipage}
    }\hspace{-15mm}
    \subfigure[Query 2/Action 5]{
        \begin{minipage}[t]{0.39\linewidth}
        \centering
        \includegraphics[width=2.2in]{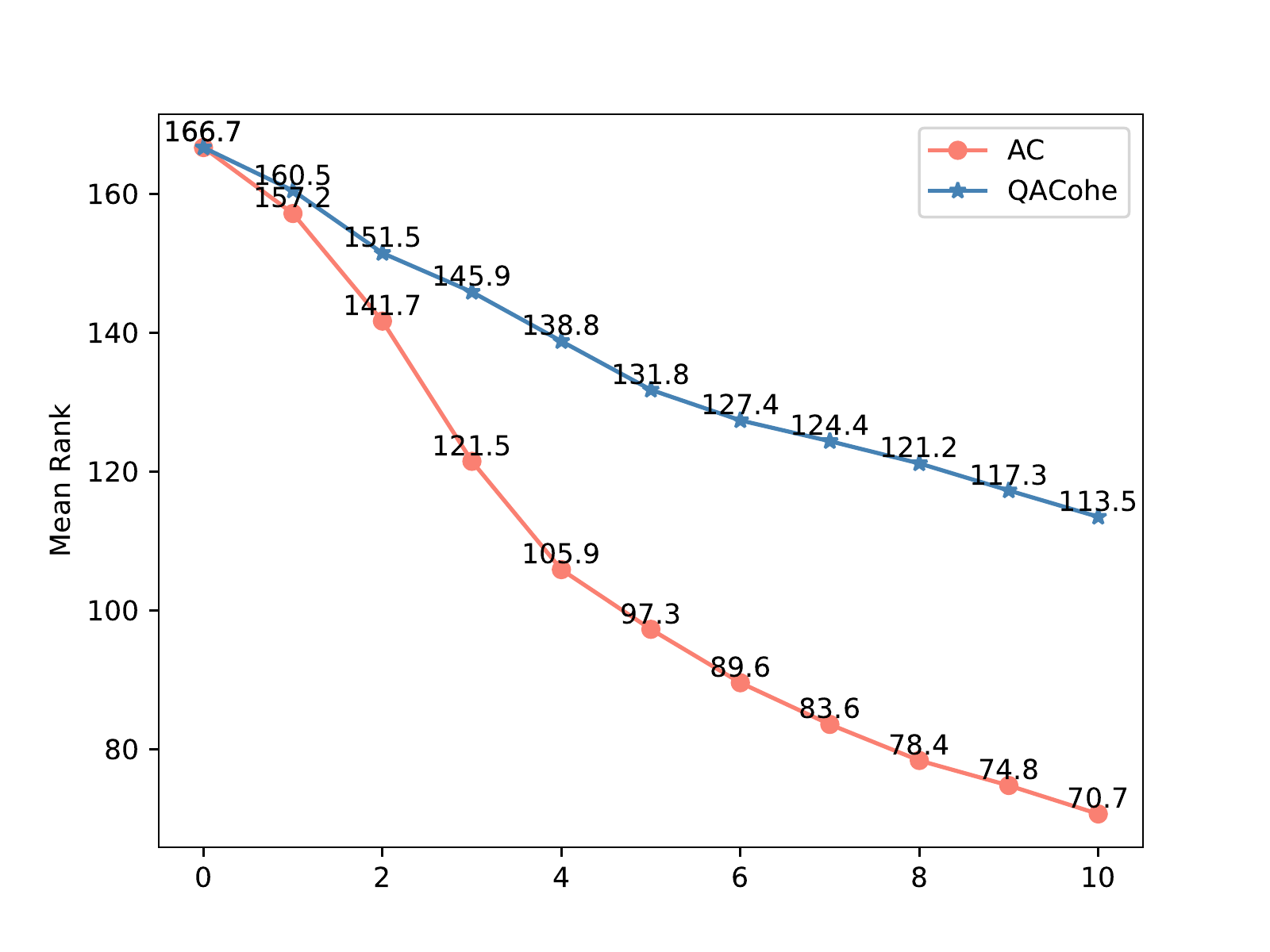}
        \vspace{-10mm}
        \caption*{}
        \end{minipage}
    }\hspace{-15mm}
    \subfigure[Query 4/Action 3]{
        \begin{minipage}[t]{0.39\linewidth}
        \centering
        \includegraphics[width=2.2in]{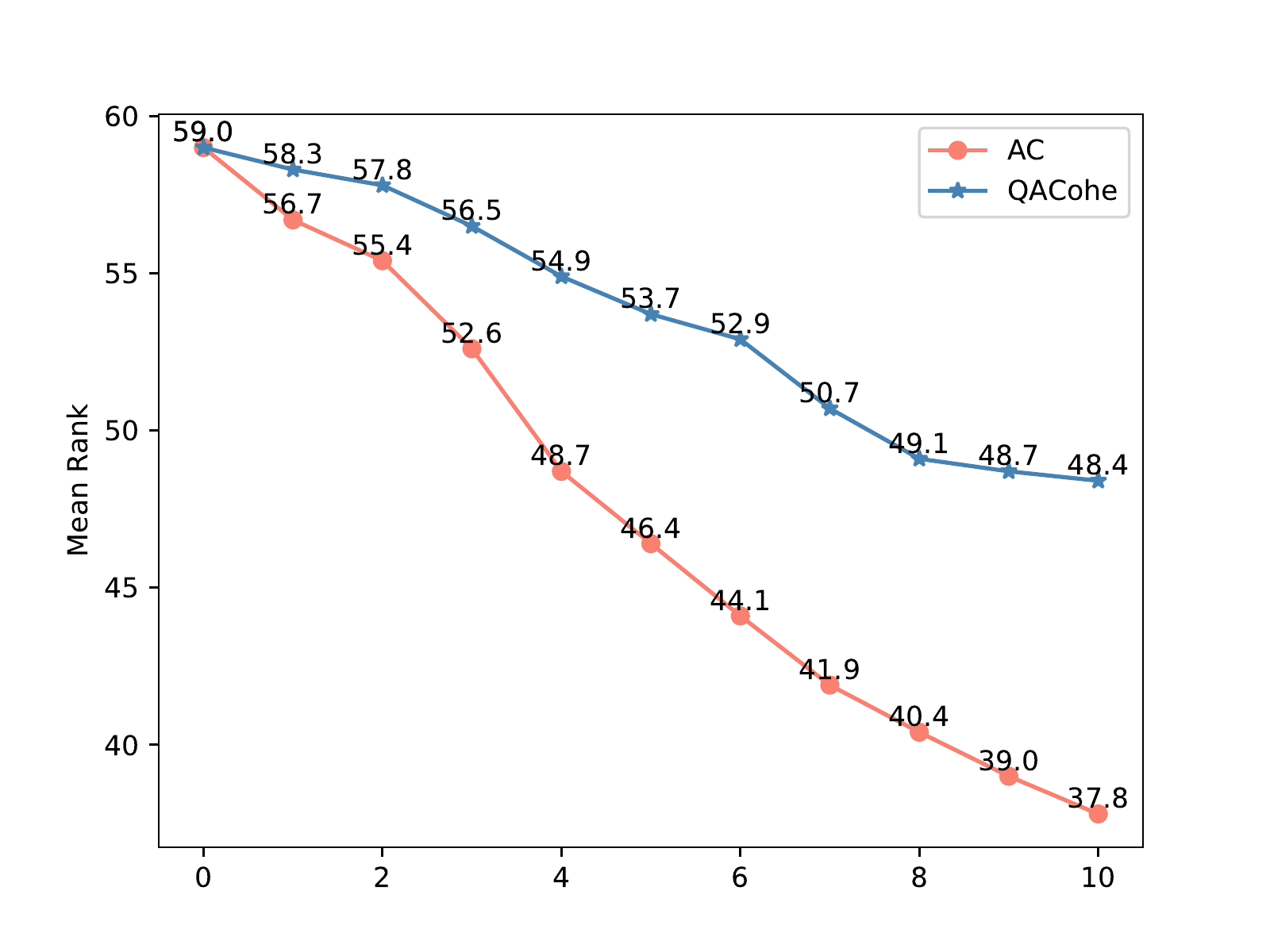}
        \vspace{-10mm}
        \caption*{}
        \end{minipage}
    }\hspace{-9mm}
    \vspace{-4mm}
    \caption{Results of AC and QACohe on Mean Rank.The horizontal axis represents the query turn.}
    \vspace{-7mm}
    \label{fig:ac_qacohe_mean}
\end{figure*}

In this section, we give a detailed comparison between AC and QACohe on R@1, R@5 and Mean Rank. Table.\ref{tab:ac_vs_qacohe} shows the comparison between AC and QACohe on R@1, R@5 and Mean Rank after 10 turns. Figure.\ref{fig:ac_qacohe_r1}, \ref{fig:ac_qacohe_r5} and \ref{fig:ac_qacohe_mean} demonstrate the performance of AC and QACohe on R@1, R@5 and Mean Rank. Meanwhile, Standard Deviation (SD) of AC, DD and WS on R@1 is 2.24, 2.18, 0.71. SD on R@5 is 1.66, 2.18, 1.41. SD on R@10 is 1.41, 2.60, 2.60. It is obvious that AC outperforms QACohe in all settings, which demonstrates that our RL-based policy is better than pre-defined policies.

\begin{table}
\begin{center}
\scriptsize
\begin{tabular}{|l|c|c|c|c|c|c|}
\hline
User & \multicolumn{2}{|c|}{AC} & \multicolumn{2}{|c|}{DD} & \multicolumn{2}{|c|}{WS} \\
\hline
 & R@1 & R@5 &R@1 & R@5 &R@1 & R@5\\
\hline\hline
Exp1& 12.0 & 36.0 & 12.0 & 40.0 & 3.0 & 16.0 \\
Nov1& 8.0 & 32.0& 8.0 & 38.0 & 2.0& 14.0 \\
Nov2& 10.0 & 32.0& 6.0& 34.0& 2.0& 14.0\\
Nov3& 6.0 & 34.0& 8.0&38.0&1.0 & 12.0\\
\hline
\end{tabular}
\end{center}
\vspace{-3mm}
\caption{Detailed performance of each user.}
\label{tab:user_performance}
\vspace{-4mm}
\end{table}



\begin{table}
    \begin{center}
    \begin{tabular}{|l|c|c|c|c|c|}
    \hline
    Method & R@1 & R@5 & MR & Q & A\\
    \hline\hline
    AC & 8.6 & 33.9 &96.0 & 1 & 10\\
    QACohe & 7.2 & 27.6 & 154.4& 1 & 10 \\
    \hline
    AC & 16.8 & 43.4& 70.7& 2 & 5 \\
    QACohe & 16.1 & 39.6 &113.5 & 2 & 5 \\
    \hline
    AC & 34.1 & 61.4 & 37.8& 4 & 3 \\
    QACohe & 33.8 & 58.7&48.4 & 4 & 3\\ 
    \hline
    \end{tabular}
    \caption{Performance of AC and QACohe on R@1, R@5 and Mean Rank. Q and A denote the number of queries and actions.}
    \label{tab:ac_vs_qacohe}
    \vspace{-8mm}
    \end{center}
\end{table}


\subsection{Visualizations}
\begin{figure*}
    \centering 
    \includegraphics[width=0.8\linewidth]{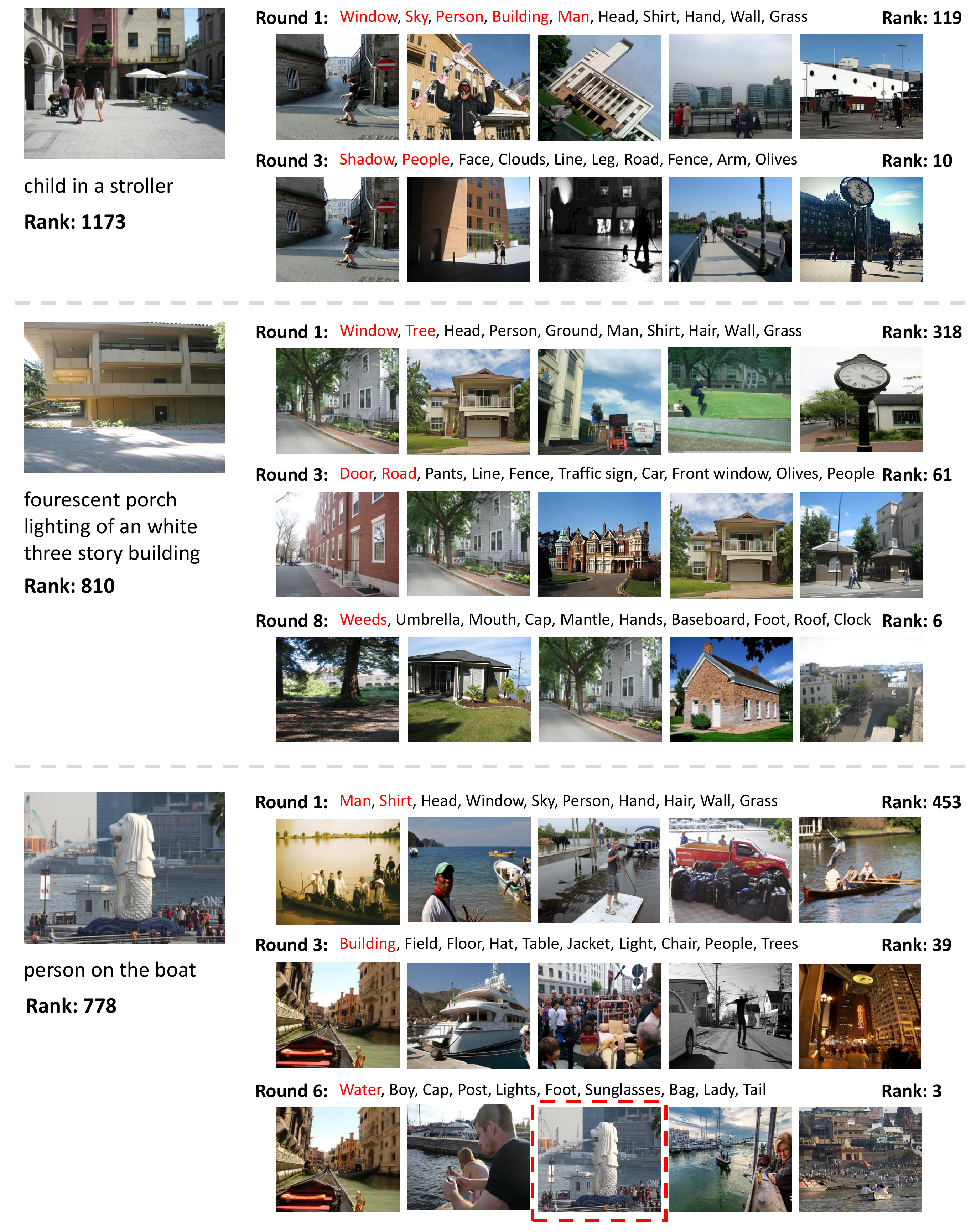}
    \caption{Visualizations of Ask\&Confirm based on SCAN with Query1/Action10.}
    \label{fig:sup_vis_q1}
\end{figure*}

\begin{figure*}
    \centering 
    \includegraphics[width=0.8\linewidth]{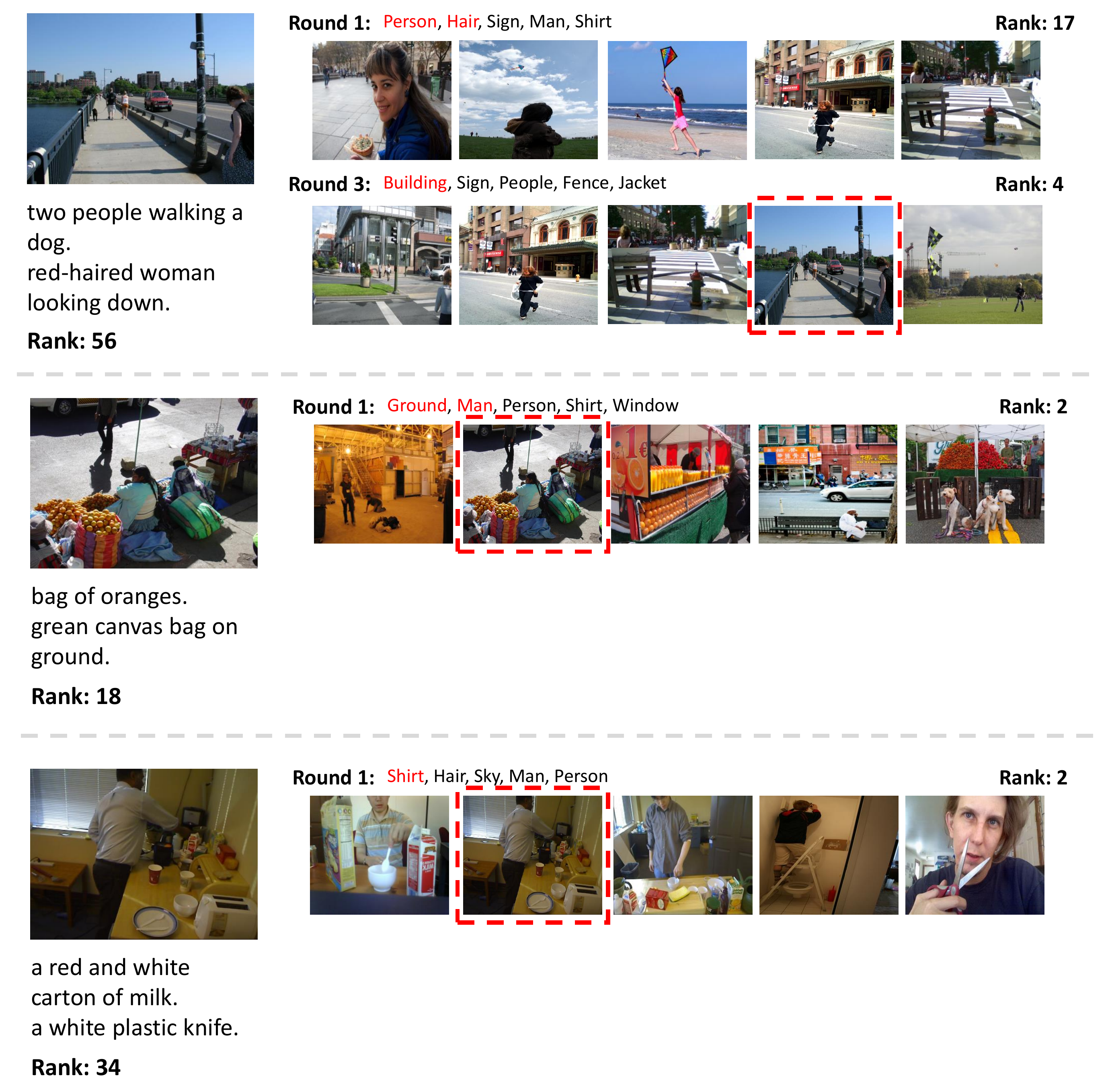}
    \caption{Visualizations of Ask\&Confirm based on SCAN with Query2/Action5.}
    \label{fig:sup_vis_q2}
\end{figure*}

\begin{figure*}
    \centering 
    \includegraphics[width=0.8\linewidth]{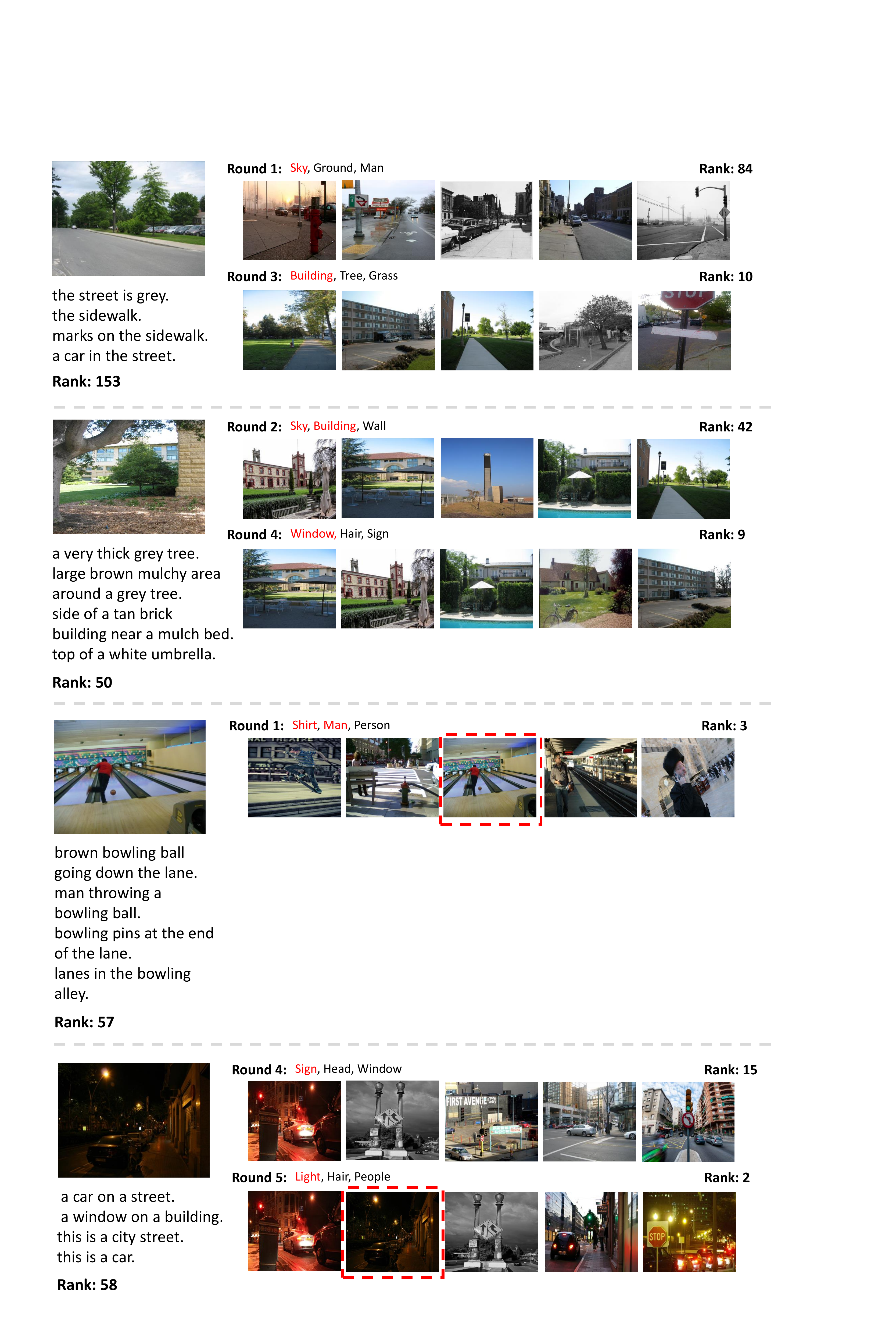}
    \caption{Visualizations of Ask\&Confirm based on SCAN with Query4/Action3.}
    \label{fig:sup_vis_q4}
\end{figure*}

In this section, we provide more visualizations of Ask\&Confirm based on SCAN~\cite{lee2018stacked} to verify the effectiveness of it. We perform Ask\&Confirm in three settings: (1) Q1/A10, (2) Q2/A5, and (3) Q4/A3. QK means K queries are given by users in the beginning, and AK means K actions are provided by an agent in each round. In detail, we visualize Ask\&Confirm with Q1/A10, Q2/A5, Q4/A3 in Figure~\ref{fig:sup_vis_q1},~\ref{fig:sup_vis_q2} and \ref{fig:sup_vis_q4}, respectively. 

\end{document}